\documentclass[runningheads]{llncs}

\usepackage{eccv}

\usepackage[T1]{fontenc}
\usepackage[utf8]{inputenc}
\usepackage[warn]{textcomp}

\usepackage{array}
\usepackage{fontawesome5}
\usepackage{enumitem} %
\usepackage{overpic} %
\usepackage{color}
\usepackage{microtype} %
\usepackage{xspace}
\usepackage{xfrac}
\usepackage{colortbl}
\usepackage{bm}
\usepackage{microtype}
\usepackage{lipsum,multicol}

\usepackage{caption}
\usepackage{subcaption}
\usepackage{wrapfig}
\usepackage{multirow}

\usepackage{graphicx}
\usepackage{amsmath}
\usepackage{amssymb}
\usepackage{booktabs}
\usepackage{adjustbox}
\usepackage{arydshln}
\usepackage[accsupp]{axessibility}  %
\usepackage{pdflscape}
\usepackage{algorithm}
\usepackage{algpseudocode}

\usepackage{siunitx}
\usepackage{soul,xcolor}
\setstcolor{red}

\usepackage{tikz}
\usepackage{pgfplots}
\pgfplotsset{compat=1.14}

\usetikzlibrary{shapes}
\usetikzlibrary{calc}
\usetikzlibrary{backgrounds}
\usetikzlibrary{positioning}
\usetikzlibrary{arrows}
\usetikzlibrary{arrows.meta}
\usetikzlibrary{decorations.text}    
\usetikzlibrary{fit}
\usetikzlibrary{spy}
\usetikzlibrary{3d}
\usetikzlibrary{positioning}
\usetikzlibrary{decorations.pathreplacing,calligraphy}

\usepackage[skins]{tcolorbox}

\definecolor{turquoise}{cmyk}{0.65,0,0.1,0.3}
\definecolor{purple}{rgb}{0.65,0,0.65}
\definecolor{darkgreen}{rgb}{0, 0.5, 0}
\definecolor{greenish}{rgb}{0.3, 0.7, 0.1}
\definecolor{lightgreen}{rgb}{0.2, 0.7, 0.2}
\definecolor{orange}{rgb}{0.7, 0.5, 0.5}
\definecolor{reddish}{rgb}{0.9, 0.3, 0.1}
\definecolor{orangeish}{rgb}{0.9, 0.6, 0.0}
\definecolor{ipadaptercolor}{rgb}{0.8, 0.7, 0.1}
\definecolor{red}{rgb}{0.8, 0.2, 0.2}
\definecolor{darkred}{rgb}{0.6, 0.1, 0.05}
\definecolor{blueish}{rgb}{0.0, 0.3, .6}
\definecolor{lightgray}{rgb}{0.7, 0.7, .7}
\definecolor{darkgray}{rgb}{0.3, 0.3, .3}
\definecolor{pink}{rgb}{1, 0, 1}
\definecolor{greyblue}{rgb}{0.25, 0.25, 1}

\definecolor{bestcol}{RGB}{254,196,79}

\definecolor{secondbestcol}{RGB}{255,247,188}

\newif\ifreview
\reviewtrue 

\newif\ifdrafting
\draftingtrue %

\ifdrafting
\newcommand{\mb}[1]{{\color{blueish}#1}} %
\newcommand{\MB}[1]{{\color{blueish}{\bf [MB: #1]}}} %
\newcommand{\Mb}[1]{\marginpar{\tiny{\textcolor{blueish}{#1}}}} %
\newcommand{\sv}[1]{{\color{darkgreen}#1}}
\newcommand{\SV}[1]{{\color{darkgreen}{\bf [SV: #1]}}}
\newcommand{\Sv}[1]{\marginpar{\tiny{\textcolor{darkgreen}{#1}}}}
\newcommand{\svs}[1]{\st{#1}}
\newcommand{\sd}[1]{{\color{greyblue}#1}}
\newcommand{\SD}[1]{{\color{greyblue}{\bf [SD: #1]}}}
\newcommand{\Sd}[1]{\marginpar{\tiny{\textcolor{greyblue}{#1}}}}
\newcommand{\mpe}[1]{{\color{purple}#1}}
\newcommand{\MP}[1]{{\color{purple}{\bf [SD: #1]}}}
\newcommand{\Mp}[1]{\marginpar{\tiny{\textcolor{purple}{#1}}}}
\else
\newcommand{\mb}[1]{} 
\newcommand{\MB}[1]{} 
\newcommand{\Mb}[1]{} 
\newcommand{\sv}[1]{} 
\newcommand{\SV}[1]{} 
\newcommand{\Sv}[1]{}
\newcommand{\svs}[1]{} 
\newcommand{\sd}[1]{} 
\newcommand{\SD}[1]{} 
\newcommand{\Sd}[1]{}
\newcommand{\mpe}[1]{} 
\newcommand{\MP}[1]{} 
\newcommand{\Mp}[1]{}
\fi

\newcommand{\vect}[1]{\bm{#1}}

\newcommand\rgt{\aftergroup\mathclose\aftergroup{\aftergroup}\right}

\makeatletter
\DeclareRobustCommand\onedot{\futurelet\@let@token\@onedot}
\def\@onedot{\ifx\@let@token.\else.\null\fi\xspace}

\def\eg{\emph{e.g}\onedot} 
\def\ie{\emph{i.e}\onedot}

\def\wrt{w.r.t\onedot}

\makeatother

\newcommand{\inlinesection}[1]{\vspace{0.05cm} \noindent {\bf #1}}

\usepackage{pifont}

\usepackage{blindtext}

\definecolor{mpurple}{RGB}{106,27,154}
\definecolor{mpurplelight}{RGB}{206,147,216}

\definecolor{mblue}{RGB}{40,53,147}
\definecolor{mbluelight}{RGB}{159,168,218}

\definecolor{mteal}{RGB}{0,105,92}
\definecolor{mteallight}{RGB}{128,203,196}

\definecolor{morangelight}{RGB}{255,171,145}

\definecolor{mgrayblue}{RGB}{55,71,79}
\definecolor{mgraybluelight}{RGB}{176,190,197}

\definecolor{mamber}{RGB}{255,143,0}
\definecolor{mamberlight}{RGB}{255,224,130}

\definecolor{mdeeporange}{RGB}{216,67,21}
\definecolor{morange}{RGB}{245,124,0}
\definecolor{myellow}{RGB}{253,216,53}

\definecolor{mgreen}{RGB}{85,139,47}
\definecolor{mgreenlight}{RGB}{174,213,129}

\definecolor{mred}{RGB}{198,40,40}
\definecolor{mredlight}{RGB}{239,154,154}

\definecolor{corange1}{RGB}{254,237,222}
\definecolor{corange2}{RGB}{253,190,133}
\definecolor{corange3}{RGB}{253,141,60}
\definecolor{corange4}{RGB}{230,85,13}
\definecolor{corange5}{RGB}{166,54,3}

\definecolor{cblue1}{RGB}{239,243,255}
\definecolor{cblue2}{RGB}{189,215,231}
\definecolor{cblue3}{RGB}{107,174,214}
\definecolor{cblue4}{RGB}{49,130,189}
\definecolor{cblue5}{RGB}{8,81,156}

\definecolor{cgray1}{RGB}{247,247,247}
\definecolor{cgray2}{RGB}{204,204,204}
\definecolor{cgray3}{RGB}{150,150,150}
\definecolor{cgray4}{RGB}{99,99,99}
\definecolor{cgray5}{RGB}{37,37,37}

\definecolor{cred1}{RGB}{254,229,217}
\definecolor{cred2}{RGB}{252,174,145}
\definecolor{cred3}{RGB}{251,106,74}
\definecolor{cred4}{RGB}{222,45,38}
\definecolor{cred5}{RGB}{165,15,21}

\definecolor{cgreen1}{RGB}{237,248,233}
\definecolor{cgreen2}{RGB}{186,228,179}
\definecolor{cgreen3}{RGB}{116,196,118}
\definecolor{cgreen4}{RGB}{49,163,84}
\definecolor{cgreen5}{RGB}{0,109,44}

\definecolor{cdiv11}{RGB}{166,97,26}
\definecolor{cdiv12}{RGB}{223,194,125}
\definecolor{cdiv13}{RGB}{245,245,245}
\definecolor{cdiv14}{RGB}{128,205,193}
\definecolor{cdiv15}{RGB}{1,133,113}

\definecolor{cdiv21}{RGB}{208,28,139}
\definecolor{cdiv22}{RGB}{241,182,218}
\definecolor{cdiv23}{RGB}{247,247,247}
\definecolor{cdiv24}{RGB}{184,225,134}
\definecolor{cdiv25}{RGB}{77,172,38}

\definecolor{cdiv31}{RGB}{230,97,1}
\definecolor{cdiv32}{RGB}{253,184,99}
\definecolor{cdiv33}{RGB}{247,247,247}
\definecolor{cdiv34}{RGB}{178,171,210}
\definecolor{cdiv35}{RGB}{94,60,153}

\definecolor{bestcol}{RGB}{255,247,188}

\newcommand{\nicenumber}[1]{\num[group-separator={,}]{#1}}
\newbox\lipsumone
\newbox\lipsumtwo
\long\def\loremlines#1{%
    \setbox\lipsumone=\vbox {%
      \lipsum
     }
   \setbox\lipsumtwo=\vsplit\lipsumone to #1\baselineskip
   \unvbox\lipsumtwo}

\usepackage{eccvabbrv}

\usepackage{graphicx}
\usepackage{booktabs}

\usepackage[accsupp]{axessibility}  %

\usepackage{hyperref}

\usepackage{orcidlink}

\usepackage[width=122mm,left=12mm,paperwidth=146mm,height=193mm,top=12mm,paperheight=217mm]{geometry}
\begin{document}

\title{Jointly Generating Multi-view Consistent PBR Textures using Collaborative Control} 
\titlerunning{Multi-view Consistent PBR Textures using Collaborative Control}

\author{Shimon Vainer, Konstantin Kutsy, Dante De Nigris, Ciara Rowles, Slava Elizarov, Simon Donné}

\authorrunning{S.~Vainer et al.}

\institute{Unity Technologies\\
Corresponding author: \email{donnesimon@gmail.com}}

\maketitle

\providecommand{\cvvimgfrac}{0.15}
\begin{figure}
\resizebox{\linewidth}{!}{
\begin{tikzpicture}[
        node distance=0cm, 
        block/.style={inner sep=0mm, draw=none, transform shape},
        prompt/.style={inner sep=0cm, text width=0.57\linewidth, align=justify, transform shape},
        overlaytext/.style={inner sep=1pt,align=center,opacity=0.8,fill=white,text opacity=1.0,rounded corners=2pt,yshift=3pt,font=\tiny},
        label/.style={inner sep=0cm, transform shape, font=\tiny},
        instruct/.style={inner sep=1pt,rounded corners=2pt, align=center,fill=white, opacity=0.8, text opacity=1.0},
    ]
    \begin{scope}[local bounding box=scenes]
    \node[block] (natural-scene) at (0,0) {\includegraphics[width=\cvvimgfrac\linewidth, height=\cvvimgfrac\linewidth]{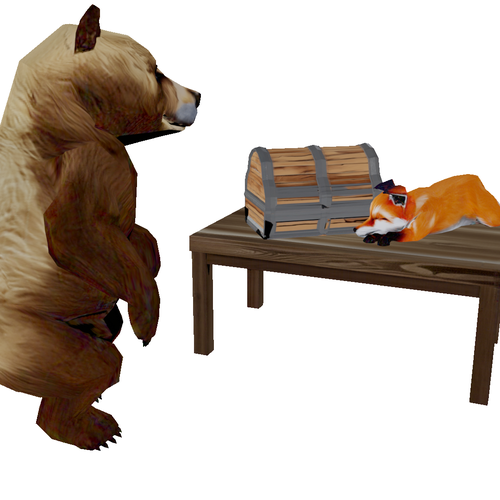}};
    \node[block, below=of natural-scene] (golden-scene) {\includegraphics[width=\cvvimgfrac\linewidth, height=\cvvimgfrac\linewidth]{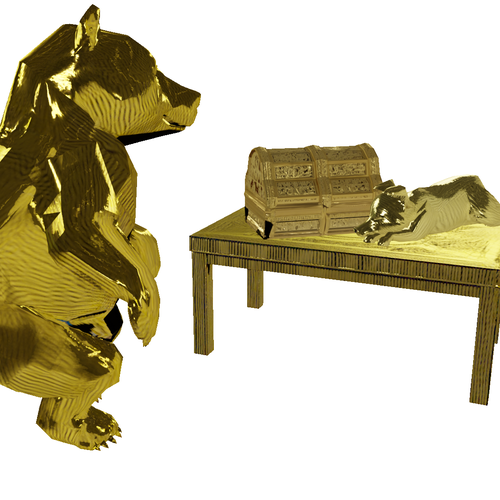}};
    \node[block, below=of golden-scene] (steampunk-scene) {\includegraphics[width=\cvvimgfrac\linewidth, height=\cvvimgfrac\linewidth]{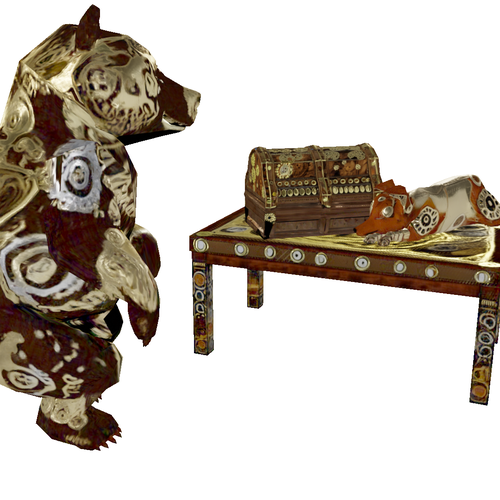}};
    \node[block, below=of steampunk-scene] (sandstone-scene) {\includegraphics[width=\cvvimgfrac\linewidth, height=\cvvimgfrac\linewidth]{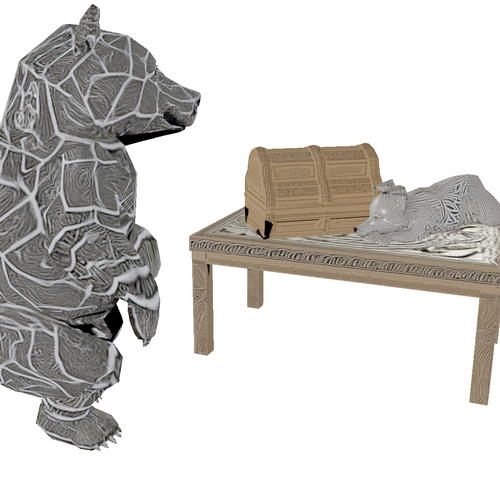}};
    \end{scope}

    \node[block] (placeholder) at (-14,0) {}; 
    \begin{scope}[xscale=-1]
    \node[block] (naked-scene) at (placeholder |- scenes) {\includegraphics[width=\cvvimgfrac\linewidth, height=\cvvimgfrac\linewidth]{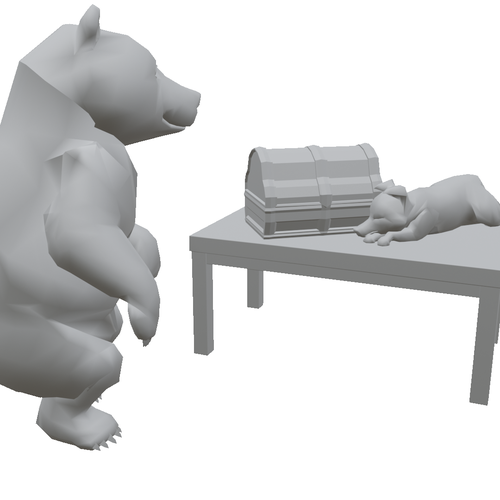}};
    \end{scope}

    \node[block] (placeholder2) at (-7,0) {};
    \node[block] (steampunk-bear) at (placeholder2 |- naked-scene) {\includegraphics[height=0.58\linewidth]{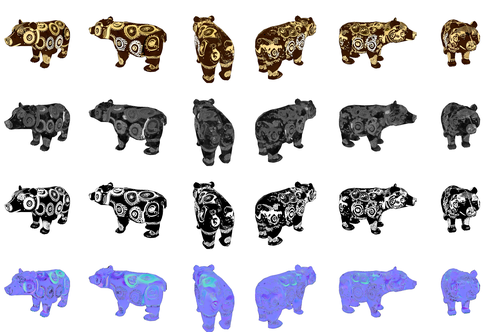}};
    \node[instruct, above=of steampunk-bear.south, yshift=0.5pt] (bear-prompt) {``Steampunk bear covered in brass cogs and gears''};
    \node[instruct, above=of natural-scene.south, yshift=1.5pt] (natural-prompt) {``Natural''};
    \node[instruct, above=of golden-scene.south, yshift=1.5pt] (golden-prompt) {``Golden''};
    \node[instruct, above=of sandstone-scene.south, yshift=1.5pt] (sandstone-prompt) {``Sandstone''};
    \draw[black!30, line width=2pt, rounded corners=3pt] (steampunk-bear.south east) rectangle (steampunk-bear.north west);
    \draw[black!30, line width=2pt, rounded corners=3pt] ($(naked-scene.south west) + (0.05, 0.05)$) rectangle ($(naked-scene.north west) + (-0.85,0.05)$);
    \draw[black!30, line width=2pt, rounded corners=3pt] ($(steampunk-scene.south west) + (-0.05, 0.05)$) rectangle ($(steampunk-scene.north west) + (0.85,0.05)$);

    \draw[black!30, line width=1pt, line cap=round] ($(naked-scene.south west) + (0.05, 0.05)$) -- (steampunk-bear.south west);
    \draw[black!30, line width=1pt, line cap=round] ($(naked-scene.north west) + (0.05, 0.05)$) -- (steampunk-bear.north west);
    \draw[black!30, line width=1pt, line cap=round] ($(steampunk-scene.south west) + (-0.05, 0.05)$) -- (steampunk-bear.south east);
    \draw[black!30, line width=1pt, line cap=round] ($(steampunk-scene.north west) + (-0.05, 0.05)$) -- (steampunk-bear.north east);

    \node[instruct, rotate=-90, xshift=-2.7cm, yshift=-5pt, font=\scriptsize] at (steampunk-bear.east) {Albedo};
    \node[instruct, rotate=-90, xshift=-0.8cm, yshift=-5pt, font=\scriptsize] at (steampunk-bear.east) {Roughness};
    \node[instruct, rotate=-90, xshift=0.9cm, yshift=-5pt, font=\scriptsize] at (steampunk-bear.east) {Metallic};
    \node[instruct, rotate=-90, xshift=2.7cm, yshift=-5pt, font=\scriptsize] at (steampunk-bear.east) {Bump map};
\end{tikzpicture}
}
\caption{We propose an end-to-end pipeline for generating graphics-ready PBR textures based only on a mesh and a text prompt. Conditioned on a single-view PBR stack, our proposed approach directly and jointly diffuses multi-view PBR images in view space. These are multi-view consistent enough that we can naively fuse them into the mesh texture.
This includes linear albedo, roughness and metallic maps, as well as normal bump maps. 
}\label{fig:teaser}
\end{figure}
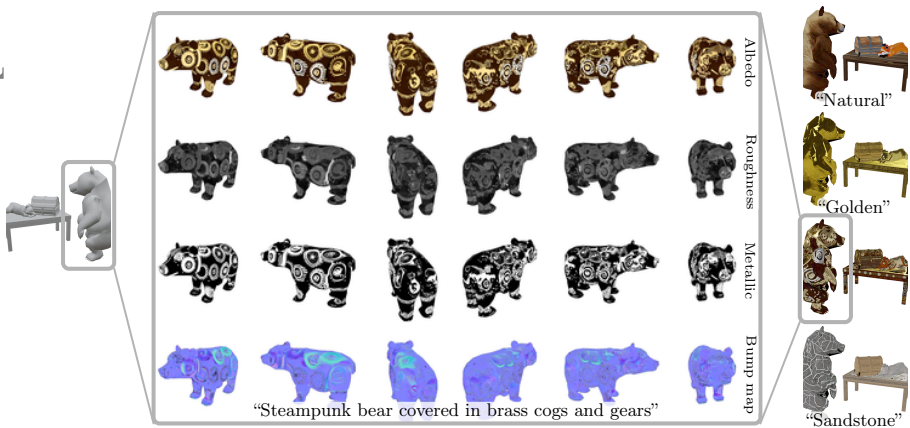

\vspace{-1cm}
\begin{abstract}
    Multi-view consistency remains a challenge for image diffusion models.
    Even within the Text-to-Texture problem, where perfect geometric correspondences are known a priori, many methods fail to yield aligned predictions across views, necessitating non-trivial fusion methods to incorporate the results onto the original mesh.
    We explore this issue for a Collaborative Control workflow specifically in PBR Text-to-Texture.
    Collaborative Control directly models PBR image probability distributions, including normal bump maps; to our knowledge, the only diffusion model to directly output full PBR stacks.
    We discuss the design decisions involved in making this model multi-view consistent, and demonstrate the effectiveness of our approach in ablation studies, as well as practical applications.
    \keywords{PBR material generation \and Texture Generation \and Multi-view \and Diffusion Models \and Collaborative Control \and Text-to-Texture}
\end{abstract}

\begin{figure}[t]
    \centering
    \includegraphics[height=0.24\linewidth]{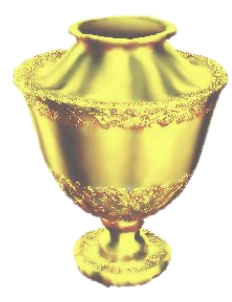}
    \hspace{1.3cm}
    \includegraphics[height=0.24\linewidth]{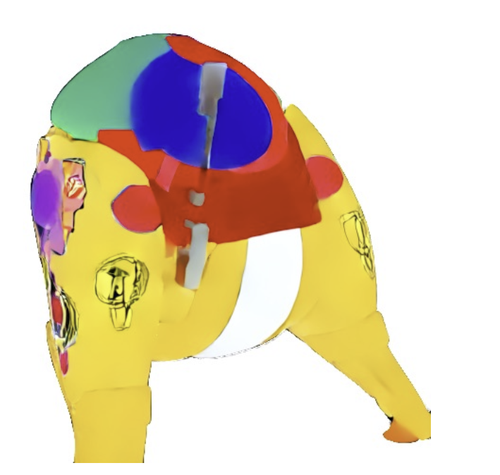}
    \hspace{13pt}
    \includegraphics[height=0.24\linewidth]{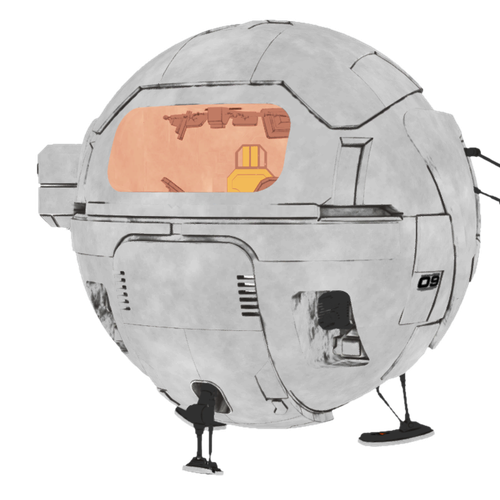}
\caption{Existing RGB-based pipelines (left) bake in lighting artifacts into the output albedo maps (taken from Fantasia3D's Figure 4~\cite{chen2023fantasia3d}). Collaborative Control (middle, from their appendix), as artist-created albedo maps (right~\cite{sphericalshipalbedo}), do not exhibit this.}\label{fig:intro}
\end{figure}

\section{Introduction}\label{sec:intro}

Although Diffusion Models~\cite{sohl2015deep, ho2020denoising, song2020denoising} have only recently garnered attention and scrutiny in the Computer Vision and Computer Graphics fields, the rate with which novel applications of these \emph{foundation models} are being developed is staggering.
Diffusion models learn the distribution of a dataset of images in order to sample novel examples from it.
In practice, they are more robust and stable to train than previous alternatives~\cite{goodfellow2020generative, karras2017progressive, karras2019style, karras2020analyzing, karras2021alias}.
They are also much easier to control: Classifier Free Guidance~\cite{Ho2022Classifier-Freearxiv} allows straightforward conditioning of the generated sample on input signals, be it text, images, or something else still.

Text-to-Texture is one such application: given an input shape (typically a mesh), the goal is to generate a texture for this shape that aligns to the input text prompt.
By now, many concurrent approaches attempt to extract such textures from an RGB-space diffusion model, by generating image samples aligned with given views of the object (be it through fine-tuning, SDS, or other methods).
But object appearance is both environment-dependent and view-dependent: the same object can appear significantly different under different lighting or from a different angle.
Modern rendering pipelines therefore represent textures as physically-based-rendering (PBR) materials that encode exactly how the surfaces react to light --- independent of both environment and viewpoint.

Inverse rendering, \ie{} estimating PBR materials from RGB measurements (typically through differentiable lighting pipelines), is ambiguous and finicky and often results in lighting artifacts and sub-optimal textures as shown in \cref{fig:intro}.
Instead, the Collaborative Control paradigm leverages the RGB model's internal state to model the PBR material distribution directly, intuiting that the model internally already disentangles the material information from the environment and view information.
As a result, it is able to directly model and sample the PBR material distribution, bypassing inverse rendering woes entirely.

However, the Collaborative Control paradigm has so far been restricted to single-view.
In this work, we illustrate how to extend the paradigm to a multi-view context
We discuss our design decisions and validate them in the experimental section, showing that our method can generate high-quality multi-view textures that are consistent enough to be fused into a texture using naive fusion techniques.

\section{Related Work}\label{sec:related}

\subsubsection{Image Generation}

Recent advances in image generation techniques have been driven by diffusion models~\cite{sohl2015deep, ho2020denoising, song2020denoising}, almost always operating in a lower-resolution latent domain that can be decoded to pixel values by a VAE~\cite{rombach2022high}.
Initial works focused around convolutional modeling with global self-attention and cross-attention to text prompts~\cite{rombach2022high,podell2023SDXL}, whereas more recent work is focused on more flexible transformer-based models, which consider images as a sequence of tokens that attend to each other and the prompt --- doing away with convolutions entirely~\cite{esser2024SD3}.
The cost of data acquisition and from-scratch training motivates us to use a pre-trained text-to-image model as a starting point.

\subsubsection{Text-to-Texture}

 considers the problem of generating appearances for an object based on a text prompt~\cite{le2023euclidreamer, youwang2023paint, zeng2023paint3d, knodt2023consistent, cao2023texfusion, chen2023text2tex, zhang2023repaint123}.
By conditioning on the known geometry (in the form of depth~\cite{le2023euclidreamer,zeng2023paint3d,knodt2023consistent,knodt2023consistent,cao2023texfusion,chen2023text2tex,zhang2023repaint123}, normals~\cite{Lu2023Direct2.5arxiv,MetaTextureGen} and/or world positions~\cite{MetaTextureGen}), such methods generate realistic appearance by reusing or fine-tuning existing text-to-image models.
However, they are typically restricted to either RGB appearance, or only albedo: metallic and roughness maps are considered too far out-of-domain for fine-tuning~\cite{liu2023unidream,MetaTextureGen}.
PBR properties for use in a graphics pipeline are obtained through inverse rendering techniques~\cite{wu2023hyperdreamer, xu2023matlaber, liu2023unidream, chen2023fantasia3d, yeh2024texturedreamer, chen2024intrinsicanything} or through various post-hoc techniques~\cite{liu2023unidream,MetaTextureGen,zeng2023paint3d}, none of which model the underlying PBR distribution properly.
To the best of our knowledge, Collaborative Control~\cite{CollabControl} is the only method to directly model the PBR image distribution for arbitrary shapes, while also predicting normal bump maps (which currently only exists for homogeneous texture patch generation~\cite{MatFuse,Sartor2023Matfusion,xu2023matlaber}), a modality even more ambiguous to extract using inverse rendering.
However, it only operates in a single view, which severely limits its applications: we extend the Collaborative Control paradigm to strongly multi-view consistent PBR generation.

\subsubsection{Multi-view Generation}

Several works~\cite{chen2023text2tex,TEXTure,InTeX,zeng2023paint3d} lift a single-view texturing model to multi-view through sequential in-painting: iteratively selecting a new viewpoint, rendering the current partial texture, and filling in missing areas as visible.
Unfortunately, these are relatively slow and not properly 3D aware: they are prone to seams and the Janus effect.
A similar issue occurs in techniques that generate multiple views independently based on a reference image and view-position encoding~\cite{Chen2023Cascade-Zero123arxiv,jeong2023nvs}.
3D Fusion of the independent views after every time-step can resolve these issues~\cite{TexPainter,cao2023texfusion,SyncMVD,liu2023syncdreamer,woo2024harmonyview,wen2024ouroboros3d}, but this curtails the diffusion process' performance as it is essentially iterative Euclidean projection.

For more holistic generation, diffusing a grid of multiple views within a single image inherently promotes multi-view consistency~\cite{wu2024unique3d,FlashTex,MetaTextureGen,shi2023mvdream,Instant-3D,shi2023zero123++}.
It is equivalent (in terms of multi-view interaction) to techniques that extend the self-attention to a global multi-view attention~\cite{long2023wonder3d,hollein2024viewdiff}.
However, just as approaches based on video models~\cite{SV3D,melas2024im3d}, these approaches still require significant effort in fusing the resulting images --- \eg{} through SDS~\cite{FlashTex}, a large-scale reconstruction model~\cite{MetaTextureGen}, or recursive fitting and re-noising of Gaussian splats~\cite{melas2024im3d}.

To achieve better multi-view consistency we require techniques that directly model cross-view correspondences.
Text-to-3D models often restrict the cross-view attention to the known epipolar lines~\cite{ye2024consistent1to3consistentimage3d,kant2024spad,wang2023mvdd,li2024era3d,Huang2023EpiDiffarxiv} or even tentative point correspondences~\cite{hu2024mvd,yang2024consistnet,ding2024bidiff}.
We leverage the knowledge of the true 3D shape by attending with each pixel only to its correspondences, as in MVDiffusion~\cite{Tang2023MVDiffusionarxiv}.

\subsubsection{Reference view}
Multi-view diffusion is much slower to iterate, so we assume a reference view to condition the generation.
This reference can be either a fixed view in the multi-view diffusion~\cite{melas2024im3d,kim2024referenceimage}, or compressed into \eg{} CLIP or DINOv2~\cite{Radford2021LearningTV,oquab2023dinov2} spaces for cross-attention~\cite{jeong2023nvs,Zheng2023Free3Darxiv}.
We leverage both: the former is paramount for accurately transferring the reference view's appearance to the final output, while the latter is crucial for ensuring that occluded areas are still efficiently textured in a coherent manner.

\begin{figure}[t]
    \centering
    \begin{tikzpicture}[spy using outlines={circle, magnification=1, connect spies, draw=black}]
        \node[inner sep=0pt] {\includegraphics[width=0.15\linewidth]{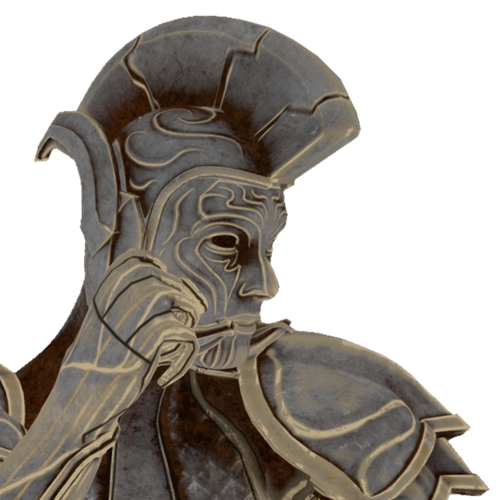}};
        \spy[size=0.30cm] on (0.3,-0.5) in node[double distance = 1pt, fill=white];
    \end{tikzpicture}
    \hspace{1.5mm}
    \begin{tikzpicture}[spy using outlines={circle, magnification=1, connect spies, draw=black}]
        \node[inner sep=0pt] {\includegraphics[width=0.15\linewidth]{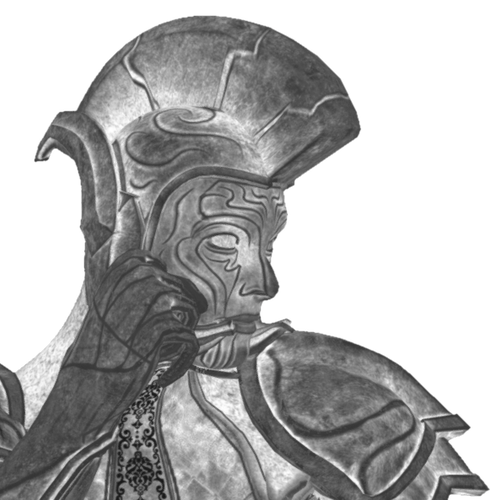}};
    \spy[size=0.30cm] on (0.3,-0.5) in node[double distance = 1pt, fill=white];
    \end{tikzpicture}
    \hspace{1.5mm}
    \begin{tikzpicture}[spy using outlines={circle, magnification=1, connect spies, draw=black}]
        \node[inner sep=0pt] {\includegraphics[width=0.15\linewidth]{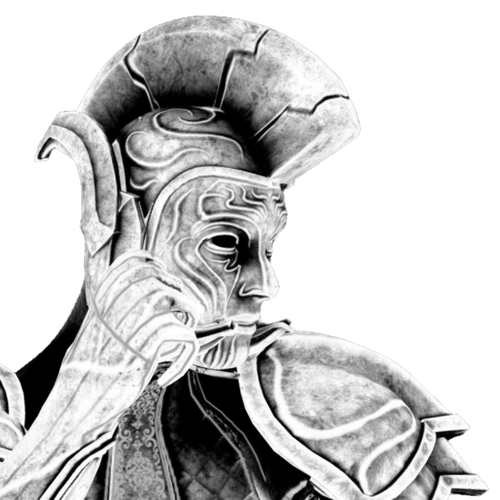}};
    \spy[size=0.30cm] on (0.3,-0.5) in node[double distance = 1pt, fill=white];
    \end{tikzpicture}
    \hspace{1.5mm}
    \begin{tikzpicture}[spy using outlines={circle, magnification=1, connect spies, draw=black}]
        \node[inner sep=0pt] {\includegraphics[width=0.15\linewidth]{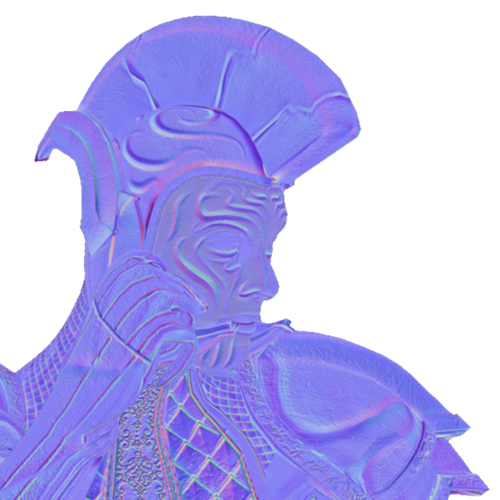}};
    \spy[size=0.30cm] on (0.3,-0.5) in node[double distance = 1pt, fill=white];
    \end{tikzpicture}
    \hspace{1.5mm}
    \begin{tikzpicture}[spy using outlines={circle, magnification=1, connect spies, draw=black}]
        \node[inner sep=0pt] {\includegraphics[width=0.15\linewidth]{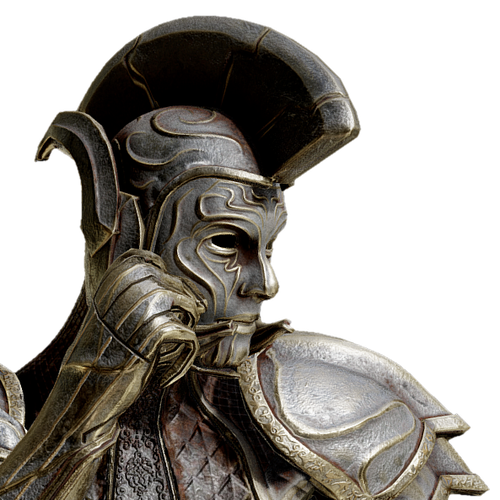}};
    \spy[size=0.30cm] on (0.3,-0.5) in node[double distance = 1pt, fill=white];
    \end{tikzpicture}

    \begin{tikzpicture}[spy using outlines={circle, magnification=1, connect spies, draw=black}]
        \node[inner sep=0pt] {\includegraphics[width=0.15\linewidth]{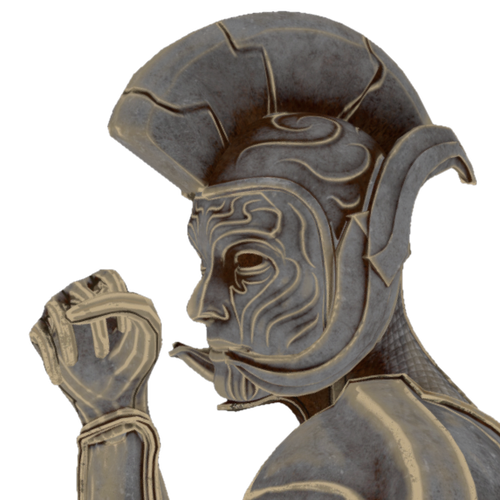}};
        \spy[size=0.30cm] on (0.2,-0.7) in node[double distance = 1pt, fill=white];
    \end{tikzpicture}
    \hspace{1.5mm}
    \begin{tikzpicture}[spy using outlines={circle, magnification=1, connect spies, draw=black}]
        \node[inner sep=0pt] {\includegraphics[width=0.15\linewidth]{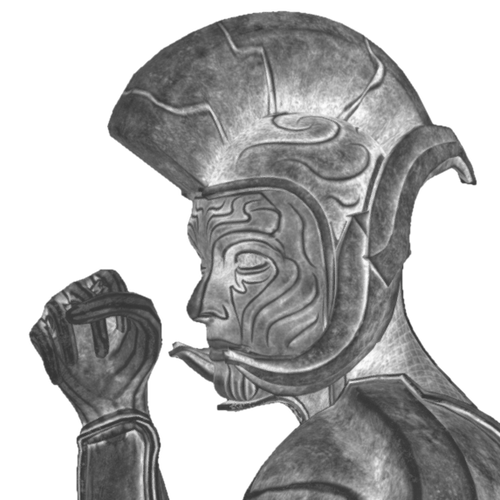}};
        \spy[size=0.30cm] on (0.2,-0.7) in node[double distance = 1pt, fill=white];
    \end{tikzpicture}
    \hspace{1.5mm}
    \begin{tikzpicture}[spy using outlines={circle, magnification=1, connect spies, draw=black}]
        \node[inner sep=0pt] {\includegraphics[width=0.15\linewidth]{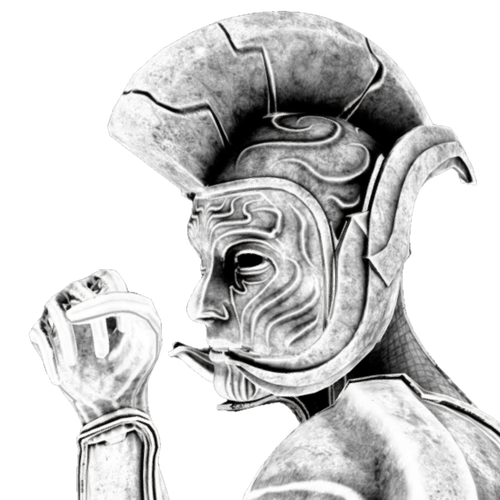}};
        \spy[size=0.30cm] on (0.2,-0.7) in node[double distance = 1pt, fill=white];
    \end{tikzpicture}
    \hspace{1.5mm}
    \begin{tikzpicture}[spy using outlines={circle, magnification=1, connect spies, draw=black}]
        \node[inner sep=0pt] {\includegraphics[width=0.15\linewidth]{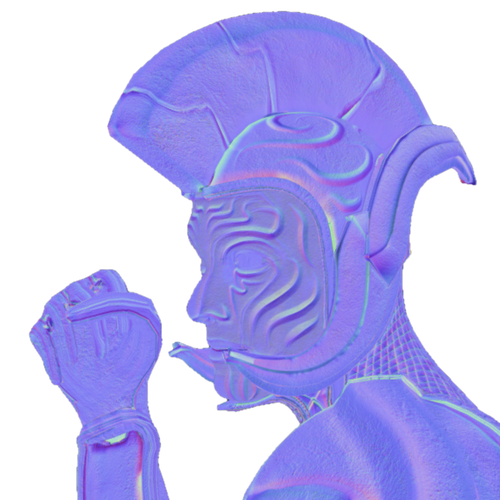}};
        \spy[size=0.30cm] on (0.2,-0.7) in node[double distance = 1pt, fill=white];
    \end{tikzpicture}
    \hspace{1.5mm}
    \begin{tikzpicture}[spy using outlines={circle, magnification=1, connect spies, draw=black}]
        \node[inner sep=0pt] {\includegraphics[width=0.15\linewidth]{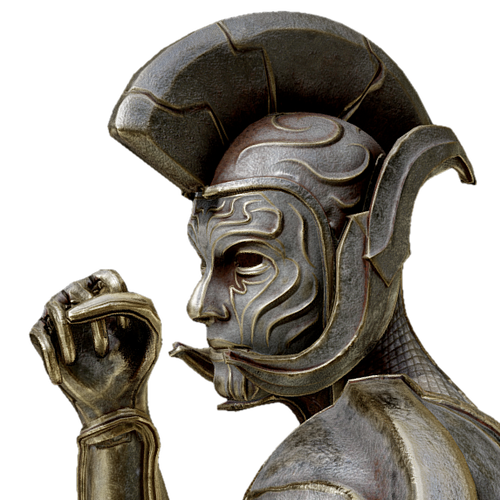}};
        \spy[size=0.30cm] on (0.2,-0.7) in node[double distance = 1pt, fill=white];
    \end{tikzpicture}
\caption{Example albedo, roughness, metallic, bump map, and rendered images from an expert artist~\cite{PBRexamplesketchfab}. Note how PBR channels are view-independent, but renders are not.}\label{fig:PBR example}
\end{figure}

\section{Method}\label{sec:method}

\subsection{Preliminaries}

\inlinesection{Diffusion models} generate samples from an image distribution $p(\vect{z})$ by learning a mapping to pure noise (typically white Gaussian noise).
They iterative sample from the conditional distribution $p(\vect{z}_{t} \vert \vect{z}_{t+1})$ to remove a small amount of noise at a time, with the \emph{time-step} $T$ varying from $T$ (pure noise) to $0$ (the target image distribution).
In practice, we wish to condition to generation to guide it with a text prompt $\mathcal{T}$ and the object geometry $\mathcal{G}$, so we learn $p(\vect{z}_{t} \vert \vect{z}_{t+1}, \mathcal{T}, \mathcal{G})$ instead.

\inlinesection{PBR materials} are a compact representation of the bidirectional reflectance distribution function (BRDF), which describes the way light is reflected from the surface of an object.
We use the Disney basecolor-metallic model~\cite{Burley2012}, a parametrization of the Cook-Torrance model that promotes physical correctness~\cite{Cook1982}; it comprises \textit{RGB albedo}, \textit{metallic}, and \textit{roughness} components.
We introduce additional local structural details through a \textit{bump normal map}, as shown in \cref{fig:PBR example}.
We use the same geometry-aligned tangent space as Collaborative Control~\cite{CollabControl} for this bump map, which is independent of the arbitrary UV space.

\inlinesection{Latent Diffusion Models} process latent images: VAE-compressed versions of the RGB- or PBR-space images~\cite{rombach2022high}.
In a multi-view setting it is paramount that the underlying data is view-independent, so that the latent values of one view are easily transferrable to other views.
PBR data is therefore better suited for multi-view diffusion than the view-dependent RGB appearance (see \cref{fig:PBR example}).

\inlinesection{Collaborative Control} models the distribution of PBR images using a diffusion model conditioned on object shape in the form of normal maps.
As no large datasets exist to train such a model from scratch, it leverages a pretrained but frozen RGB diffusion model, tightly linked to the PBR model.
This RGB-PBR model pair models the joint distribution of PBR images and their fixed-environment renders (compared to approaches like SDS~\cite{poole2022dreamfusion} which are inherently mode-seeking): the PBR model both controls the RGB generation process towards render-like images and extracts relevant information to generate its own PBR output.
The authors show that this paradigm is able to extract much knowledge from the frozen model with only a small amount of training data, while still retaining the generalizability, quality and diversity of the base RGB model.

\subsection{Architecture}

We now describe how to lift the Collaborative Control paradigm to a multi-view setting, conditioned on both object shape and a reference view's appearance: the goal is to generate PBR images that are pixel-wise consistent (up to variance in the VAE decoder) and can be naively merged into a texture representation.
We freeze the reference image and will use it to guide the multi-view generation process~\cite{melas2024im3d,kim2024referenceimage}.
An overview of our architecture is shown in \cref{fig:architecture overview}, contrasted against both the vanilla RGB architecture~\cite{rombach2022high} and Collaborative Control's single-view architecture~\cite{CollabControl}.
We discuss how to condition on the reference view, how to leverage the known correspondences from the fixed shape, and how to fuse the multi-view PBR images into a single texture map.

\subsubsection{Notations}

We take as input a set of noisy latent RGB and PBR images pairs $\left\{\vec{z}_{\textrm{RGB}, i, t}\right\}$ and $\left\{\vec{z}_{\textrm{PBR}, i, t}\right\}$ ($i=0$ is the reference view) with known camera poses $\left\{\vec{c}_{i}\right\}$ and intrinsics $\left\{K_{i}\right\}$.
The denoising is conditioned on a text prompt $\mathcal{T}$ and per-view encoded normal maps $\left\{\vec{z}_{\textrm{n}, i}\right\}$.
Where relevant, we precondition the new layer inputs by adding a Plücker coordinate embedding of the camera rays~\cite{LFNPlucker}.

\subsubsection{Cross attention to the reference image's hidden states}

Similar to~\cite{melas2024im3d,kim2024referenceimage}, we process the reference view with its own diffusion process (unperturbed by noise, with a fixed timestep $0$) and cross-attend from every view to the reference internal states at the same layer.

\subsubsection{Reference DINOv2 feature attention}

We also cross attend to the DINOv2 features~\cite{oquab2023dinov2} (after a linear projection inspired by ~\cite{brooks2023instructpix2pix}) of the reference RGB image.
This promotes global semantic reasoning and provides further semantic guidance to the new views~\cite{jeong2023nvs,Zheng2023Free3Darxiv}.

\begin{figure}[t]
    \centering
    \begin{subfigure}[b]{0.43\linewidth}
        \centering
        \resizebox{\linewidth}{!}{
    \begin{tikzpicture}[
        node distance=1cm, 
        font={\fontsize{28pt}{28}\selectfont},
        label/.style={inner sep=0.4cm, rounded corners=8pt, line width=3pt, transform shape},
        state/.style={label, draw=none, fill=none},
        attention/.style={label, text width=6.5cm, align=center, fill=white},
        residual/.style={label, fill=white, rounded corners=20pt},
        rgb/.style={draw=blueish},
        pbr/.style={draw=reddish},
        multiview/.style={draw=greenish},
        hidden/.style={draw=none, fill=none, text opacity=0},
        arrow/.style={-stealth, line width=3pt}
    ]

    \begin{scope}[shift={(0,0)},local bounding box=rgbscope]
        \node[state] (rgb-state-in) at (0, 0) {$\vec{z}_{\textrm{RGB}, t+1}$};
        \node[attention, rgb, below=of rgb-state-in] (rgb-self-attention) {Self-attention};
        \node[hidden, left=of rgb-self-attention] (rgb-self-residual-phantom) {};
        \node[residual, rgb, below=of rgb-self-attention] (rgb-self-attention-plus) {+};
        \draw[arrow] (rgb-state-in) -| (rgb-self-residual-phantom.center) |- (rgb-self-attention-plus);
        \draw[arrow] (rgb-state-in) -- (rgb-self-attention);
        \draw[arrow] (rgb-self-attention) -- (rgb-self-attention-plus);

        \node[attention, multiview, below=of rgb-self-attention-plus] (rgb-mv-attention) {Multi-view communication};
        \draw[arrow] (rgb-self-attention-plus) -- (rgb-mv-attention);

        \node[state, below=of rgb-mv-attention] (rgb-state-self) {hidden state};
        \draw[arrow] (rgb-mv-attention) -- (rgb-state-self);

        \node[attention, rgb, below=of rgb-state-self] (rgb-pbr-attention) {\phantom{Cross-attention}}; %
        \node[hidden, left=of rgb-pbr-attention] (rgb-pbr-residual-phantom) {};
        \node[residual, rgb, below=of rgb-pbr-attention] (rgb-pbr-attention-plus) {+};
        \draw[arrow] (rgb-state-self) -| (rgb-pbr-residual-phantom.center) |- (rgb-pbr-attention-plus);
        \draw[arrow] (rgb-state-self) -- (rgb-pbr-attention);
        \draw[arrow] (rgb-pbr-attention) -- (rgb-pbr-attention-plus);

        \node[state, below=of rgb-pbr-attention-plus] (rgb-state-pbr) {hidden state};
        \draw[arrow] (rgb-pbr-attention-plus) -- (rgb-state-pbr);

        \node[attention, rgb, below=of rgb-state-pbr] (rgb-prompt-attention) {Cross-attention};
        \node[hidden, left=of rgb-prompt-attention] (rgb-prompt-residual-phantom) {};
        \node[residual, rgb, below=of rgb-prompt-attention] (rgb-prompt-attention-plus) {+};
        \draw[arrow] (rgb-state-pbr) -| (rgb-prompt-residual-phantom.center) |- (rgb-prompt-attention-plus);
        \draw[arrow] (rgb-state-pbr) -- (rgb-prompt-attention);
        \draw[arrow] (rgb-prompt-attention) -- (rgb-prompt-attention-plus);

        \node[state, below=of rgb-prompt-attention-plus] (rgb-state-out) {$\vec{z}_{\textrm{RGB}, t}$};
        \draw[arrow] (rgb-prompt-attention-plus) -- (rgb-state-out);
    \end{scope}

    \begin{scope}[shift={(12,0)},local bounding box=pbrscope]
        \node[state] (pbr-state-in) at (0, 0) {$\vec{z}_{\textrm{PBR}, t+1}$};
        \node[attention, pbr, below=of pbr-state-in] (pbr-self-attention) {Self-attention};
        \node[hidden, right=of pbr-self-attention] (pbr-self-residual-phantom) {};
        \node[residual, pbr, below=of pbr-self-attention] (pbr-self-attention-plus) {+};
        \draw[arrow] (pbr-state-in) -| (pbr-self-residual-phantom.center) |- (pbr-self-attention-plus);
        \draw[arrow] (pbr-state-in) -- (pbr-self-attention);
        \draw[arrow] (pbr-self-attention) -- (pbr-self-attention-plus);

        \node[attention, multiview, below=of pbr-self-attention-plus] (pbr-mv-attention) {Multi-view communication};
        \draw[arrow] (pbr-self-attention-plus) -- (pbr-mv-attention);

        \node[state, below=of pbr-mv-attention] (pbr-state-self) {hidden state};
        \draw[arrow] (pbr-mv-attention) -- (pbr-state-self);

        \node[attention, pbr, below=of pbr-state-self] (pbr-rgb-attention) {\phantom{Cross-attention}}; %
        \node[hidden, right=of pbr-rgb-attention] (pbr-rgb-residual-phantom) {};
        \node[residual, pbr, below=of pbr-rgb-attention] (pbr-rgb-attention-plus) {+};
        \draw[arrow] (pbr-state-self) -| (pbr-rgb-residual-phantom.center) |- (pbr-rgb-attention-plus);
        \draw[arrow] (pbr-state-self) -- (pbr-rgb-attention);
        \draw[arrow] (pbr-rgb-attention) -- (pbr-rgb-attention-plus);

        \node[state, hidden, below=of pbr-rgb-attention-plus] (pbr-state-rgb) {hidden state};
        \node[attention, hidden,  below=of pbr-state-rgb] (pbr-prompt-attention) {Cross-attention};
        \node[residual, hidden, below=of pbr-prompt-attention] (pbr-prompt-attention-plus) {+};

        \node[state, below=of pbr-prompt-attention-plus] (pbr-state-out) {$\vec{z}_{\textrm{PBR}, t}$};
        \draw[arrow] (pbr-rgb-attention-plus) -- (pbr-state-out);
    \end{scope}

    \begin{scope}[on background layer,local bounding box=rgbbox]
        \fill[fill=blueish!20, rounded corners=8pt] 
        ($(rgbscope.north west)+(-0.75, 1.0)$) -- 
        ($(rgbscope.north east)+( 0.75, 1.0)$) -- 
        ($(rgbscope.south east)+( 0.75,-1.0)$) -- 
        ($(rgbscope.south west)+(-0.75,-1.0)$) -- cycle;
    \end{scope}

    \begin{scope}[on background layer,local bounding box=pbrbox]
        \fill[fill=reddish!20, rounded corners=8pt] 
        ($(pbrscope.north west)+(-0.75, 1.0)$) -- 
        ($(pbrscope.north east)+( 0.75, 1.0)$) -- 
        ($(pbrscope.south east)+( 0.75,-1.0)$) -- 
        ($(pbrscope.south west)+(-0.75,-1.0)$) -- cycle;
    \end{scope}

    \begin{scope}[local bounding box=cross-domain-box]
    \fill[fill=white, draw=reddish, rounded corners=8pt, line width=3pt] 
    ($(rgb-pbr-attention.north west)$) -- 
    ($(pbr-rgb-attention.north east)$) -- 
    ($(pbr-rgb-attention.south east)$) -- 
    ($(rgb-pbr-attention.south west)$) -- cycle;
    \end{scope}
    \node[state] (cross-domain) at (cross-domain-box.center) {Cross-domain communication};

    \node[state,draw=none,text=blueish,anchor=center,fill=white!50,inner sep=0.3cm] (RGB-labelt) at (rgbbox.north) {RGB Diffusion};
    \node[state,draw=none,text=reddish,anchor=center,fill=white!50,inner sep=0.3cm] (PBR-labelt) at (pbrbox.north) {PBR Diffusion};

    \end{tikzpicture}
}
\vspace*{-1cm}
        \caption{}
    \end{subfigure}
    \hspace{1.5cm}
    \begin{subfigure}[b]{0.39\linewidth}
        \centering
        \resizebox{\linewidth}{!}{
    \begin{tikzpicture}[
        node distance=1cm, 
        font={\fontsize{28pt}{28}\selectfont},
        label/.style={inner sep=0.4cm, rounded corners=8pt, line width=3pt, transform shape},
        state/.style={label, draw=none, fill=none},
        attention/.style={label, text width=6.5cm, align=center, fill=white},
        residual/.style={label, fill=white, rounded corners=20pt},
        rgb/.style={draw=blueish},
        pbr/.style={draw=reddish},
        multiview/.style={draw=greenish},
        hidden/.style={draw=none, fill=none, text opacity=0},
        arrow/.style={-stealth, line width=3pt}
    ]

    \begin{scope}[shift={(0,0)},local bounding box=mvscope]
        \node[state] (mv-state-in) at (0, 0) {hidden state};
        \node[residual, multiview, below=of mv-state-in] (mv-point-plucker-in) {+};
        \draw[arrow] (mv-state-in) -- (mv-point-plucker-in);
        \node[attention, multiview, below=of mv-point-plucker-in] (mv-point-plucker) {\phantom{Cpd}\\\phantom{pt}};
        \node[hidden,left=of mv-point-plucker] (mv-point-plucker-phantom) {};
        \draw[arrow] (mv-point-plucker-in) -- (mv-point-plucker);
        \node[residual, multiview, below=of mv-point-plucker] (mv-point-plucker-plus) {+};
        \draw[arrow] (mv-point-plucker) -- (mv-point-plucker-plus);
        \draw[arrow] (mv-state-in) -| (mv-point-plucker-phantom.center) |- (mv-point-plucker-plus);

        \node[state, below=of mv-point-plucker-plus] (mv-state-refstate) {hidden state};
        \draw[arrow] (mv-point-plucker-plus) -- (mv-state-refstate);

        \node[residual, multiview, below=of mv-state-refstate] (mv-refstate-in) {+};
        \draw[arrow] (mv-state-refstate) -- (mv-refstate-in);
        \node[attention, multiview, below=of mv-refstate-in] (mv-refstate) {Cross attention};
        \node[hidden,left=of mv-refstate] (mv-refstate-phantom) {};
        \draw[arrow] (mv-refstate-in) -- (mv-refstate);
        \node[residual, multiview, below=of mv-refstate] (mv-refstate-plus) {+};
        \draw[arrow] (mv-state-refstate) -| (mv-refstate-phantom.center) |- (mv-refstate-plus);
        \draw[arrow] (mv-refstate) -- (mv-refstate-plus);

        \node[state, below=of mv-refstate-plus] (mv-state-refdino) {hidden state};
        \draw[arrow] (mv-refstate-plus) -- (mv-state-refdino);

        \node[residual, multiview, below=of mv-state-refdino] (mv-refdino-in) {+};
        \draw[arrow] (mv-state-refdino) -- (mv-refdino-in);
        \node[attention, multiview, below=of mv-refdino-in] (mv-refdino) {Cross attention};
        \node[hidden,left=of mv-refdino] (mv-refdino-phantom) {};
        \draw[arrow] (mv-refdino-in) -- (mv-refdino);
        \node[residual, multiview, below=of mv-refdino] (mv-refdino-plus) {+};
        \draw[arrow] (mv-refdino) -- (mv-refdino-plus);
        \draw[arrow] (mv-state-refdino) -| (mv-refdino-phantom.center) |- (mv-refdino-plus);

        \node[state, below=of mv-refdino-plus] (mv-state-out) {hidden state};
        \draw[arrow] (mv-refdino-plus) -- (mv-state-out);
    \end{scope}

    \begin{scope}[on background layer,local bounding box=mvbox]
        \fill[fill=greenish!20!white, rounded corners=8pt] 
        ($(mvscope.north west)+(-1, 1.5)$) -- 
        ($(mvscope.north east)+( 1, 1.5)$) -- 
        ($(mvscope.south east)+( 1,-1.5)$) -- 
        ($(mvscope.south west)+(-1,-1.5)$) -- cycle;
    \end{scope}

    \node[state,align=center,text width=6cm] (plucker-embedding) at ($(mv-state-in) + (16,0)$) {Plücker ray embedding};
    \draw[arrow] (plucker-embedding) |- (mv-point-plucker-in);
    \draw[arrow] (plucker-embedding) |- (mv-refstate-in);
    \draw[arrow] (plucker-embedding) |- (mv-refdino-in);
 
    \node[attention, multiview, anchor=north west, text width=10cm] (pointwise-attention-long) at (mv-point-plucker.north west) {Correspondence point-wise attention};

    \begin{scope}[shift={(0,0)},local bounding box=refvscope]
        \node[state,right=4cm of mv-state-refstate] (refvrefstatephantom) {\phantom{hidden state}};%
    \end{scope}
    \begin{scope}[local bounding box=refvbox]
        \fill[fill=greenish!25, rounded corners=0pt,opacity=0.8]
        ($(refvscope.north west)+(-1, 0.5)$) -- 
        ($(refvscope.north east)+( 1, 0.5)$) -- 
        ($(refvscope.south east)+( 1,-0.5)$) -- 
        ($(refvscope.south west)+(-1,-0.5)$) -- cycle;
    \end{scope}
    \begin{scope}[local bounding box=refvshading]
        \shade[shading=axis,bottom color=greenish!25, top color=white, rounded corners=0pt,draw=none,opacity=0.8]
        ($(refvscope.north west)+(-1, 3.5)$) -- 
        ($(refvscope.north east)+( 1, 3.5)$) -- 
        ($(refvscope.north east)+( 1, 0.5)$) -- 
        ($(refvscope.north west)+(-1, 0.5)$) -- cycle;
        \shade[shading=axis,top color=greenish!25, bottom color=white, rounded corners=0pt,draw=none,opacity=0.8]
        ($(refvscope.south west)+(-1, -3.5)$) -- 
        ($(refvscope.south east)+( 1, -3.5)$) -- 
        ($(refvscope.south east)+( 1, -0.5)$) -- 
        ($(refvscope.south west)+(-1, -0.5)$) -- cycle;
    \end{scope}
    \node[state,draw=none,text=greenish,anchor=center,fill=white!50,inner sep=0.3cm] (refv-labelt) at (refvbox.north) {Reference view};
    \node[state] (refvrefstate) at (refvrefstatephantom) {hidden state};
    \draw[arrow] (refvrefstate) |- (mv-refstate);

    \node[state, right=2cm of mv-refdino,text width=8.5cm,align=center] (refvdinostate) {Reference view\\DINOv2 features};
    \draw[arrow] (refvdinostate) -- (mv-refdino);

    \end{tikzpicture}
}
\vspace*{-1cm}
        \caption{}\label{fig:multiview attention}
    \end{subfigure}
    \caption{(a) Structure of the existing attention blocks in vanilla \textcolor{blueish}{Stable Diffusion 2.1} and \textcolor{reddish}{Collaborative Control}, as well as the \textcolor{greenish}{multi-view communication} we introduce. (b) The components of the multi-view communication block.}\label{fig:architecture overview}
\end{figure}
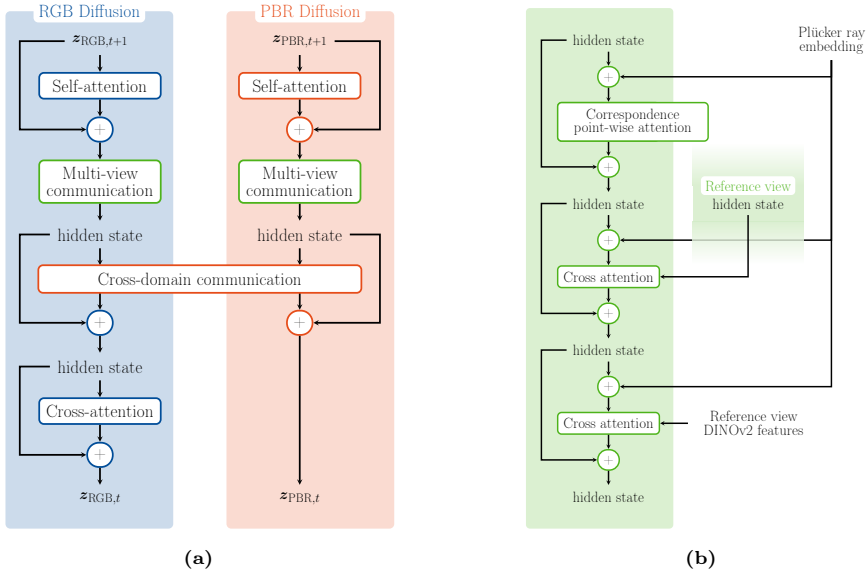

\subsubsection{Occlusion-aware point correspondence attention}

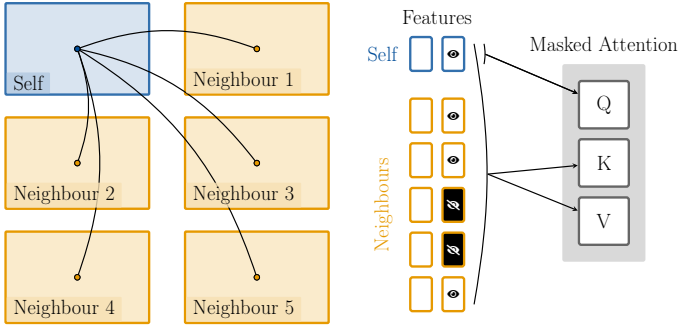
\begin{figure}[t]
    \centering
    \newlength{\cvlabelpadding}\setlength{\cvlabelpadding}{1mm}%
\resizebox{0.75\linewidth}{!}{
    \begin{tikzpicture}[
        node distance=0.25cm, 
        font={\fontsize{16pt}{16}\selectfont},
        htext/.style={inner sep=0.1cm,rounded corners=2pt,line width=2pt,},
        label/.style={htext,opacity=1.0,text opacity=1.0,rounded corners=1pt,anchor=south west},
        block/.style={rounded corners=1pt, line width=2pt, transform shape},
        selfblock/.style={block,draw=blueish!80, fill=blueish!15},
        neighbourblock/.style={block,draw=orangeish, fill=orangeish!20},
        hidden/.style={draw=none, fill=none, text opacity=0},
        feature/.style={htext,text width=0.45cm,minimum height=1.0cm,draw=black,align=center,font=\small},
        attention/.style={htext,fill=white,text width=1.3cm,minimum height=1.4cm,draw=black!60,align=center},
        brace/.style={thick,decorate,decoration={calligraphic brace, amplitude=7pt,raise=0.5ex}}
    ]
    
    \begin{scope}[scale=0.27, local bounding box=blocks]
        \begin{scope}[shift={(0,0)},local bounding box=blockTL]
        \fill[selfblock]
        (0,0) -- (16,0) -- (16,-10) -- (0,-10) -- cycle;
        \end{scope}

        \begin{scope}[shift={(20,0)},local bounding box=blockTM]
        \fill[neighbourblock]
        (0,0) -- (16,0) -- (16,-10) -- (0,-10) -- cycle;
        \end{scope}

        \begin{scope}[shift={(0,-12.5)},local bounding box=blockTR]
        \fill[neighbourblock]
        (0,0) -- (16,0) -- (16,-10) -- (0,-10) -- cycle;
        \end{scope}

        \begin{scope}[shift={(20,-12.5)},local bounding box=blockBL]
        \fill[neighbourblock]
        (0,0) -- (16,0) -- (16,-10) -- (0,-10) -- cycle;
        \end{scope}

        \begin{scope}[shift={(0,-25)},local bounding box=blockBM]
        \fill[neighbourblock]
        (0,0) -- (16,0) -- (16,-10) -- (0,-10) -- cycle;
        \end{scope}

        \begin{scope}[shift={(20,-25)},local bounding box=blockBR]
        \fill[neighbourblock]
        (0,0) -- (16,0) -- (16,-10) -- (0,-10) -- cycle;
        \end{scope}

        \filldraw[fill=blueish] (blockTL.center) circle (0.3cm);
        \filldraw[fill=orangeish] (blockTM.center) circle (0.3cm);
        \filldraw[fill=orangeish] (blockTR.center) circle (0.3cm);
        \filldraw[fill=orangeish] (blockBL.center) circle (0.3cm);
        \filldraw[fill=orangeish] (blockBM.center) circle (0.3cm);
        \filldraw[fill=orangeish] (blockBR.center) circle (0.3cm);
        \draw[line width=1pt] (blockTL.center) to[bend left=20] (blockTM.center);
        \draw[line width=1pt] (blockTL.center) to[bend left=20] (blockTR.center);
        \draw[line width=1pt] (blockTL.center) to[bend left=20] (blockBL.center);
        \draw[line width=1pt] (blockTL.center) to[bend left=20] (blockBM.center);
        \draw[line width=1pt] (blockTL.center) to[bend left=20] (blockBR.center);
        \filldraw[fill=blueish] (blockTL.center) circle (0.3cm);
        \filldraw[fill=orangeish] (blockTM.center) circle (0.3cm);
        \filldraw[fill=orangeish] (blockTR.center) circle (0.3cm);
        \filldraw[fill=orangeish] (blockBL.center) circle (0.3cm);
        \filldraw[fill=orangeish] (blockBM.center) circle (0.3cm);
        \filldraw[fill=orangeish] (blockBR.center) circle (0.3cm);

        \node[label,fill=blueish!30, opacity=0.8, text opacity=1.0] at (blockTL.south west) {\hspace*{\cvlabelpadding}Self\hspace*{\cvlabelpadding}};
        \node[label,fill=orangeish!30, opacity=0.8, text opacity=1.0] at (blockTM.south west) {\hspace*{\cvlabelpadding}Neighbour 1\hspace*{\cvlabelpadding}};
        \node[label,fill=orangeish!30, opacity=0.8, text opacity=1.0] at (blockTR.south west) {\hspace*{\cvlabelpadding}Neighbour 2\hspace*{\cvlabelpadding}};
        \node[label,fill=orangeish!30, opacity=0.8, text opacity=1.0] at (blockBL.south west) {\hspace*{\cvlabelpadding}Neighbour 3\hspace*{\cvlabelpadding}};
        \node[label,fill=orangeish!30, opacity=0.8, text opacity=1.0] at (blockBM.south west) {\hspace*{\cvlabelpadding}Neighbour 4\hspace*{\cvlabelpadding}};
        \node[label,fill=orangeish!30, opacity=0.8, text opacity=1.0] at (blockBR.south west) {\hspace*{\cvlabelpadding}Neighbour 5\hspace*{\cvlabelpadding}};
    \end{scope}

    \begin{scope}[shift={(12.5,-1.5)}, local bounding box=all-data]
        \begin{scope}[local bounding box=self-data]
            \node[feature, draw=blueish!80] (self-feature) at (0,0) {};
            \node[feature, draw=blueish!80, fill=white, right=of self-feature] (self-occlusion) {\faEye};
        \end{scope}

        \begin{scope}[local bounding box=neighbours-data]
            \node[feature, draw=orangeish, below=0.75cm of self-feature] (neighbour1-feature) {};
            \node[feature, draw=orangeish, below=0.75cm of self-occlusion, fill=white] (neighbour1-occlusion) {\faEye};
            \node[feature, draw=orangeish, below=of neighbour1-feature] (neighbour2-feature) {};
            \node[feature, draw=orangeish, below=of neighbour1-occlusion, fill=white] (neighbour2-occlusion) {\faEye};
            \node[feature, draw=orangeish, below=of neighbour2-feature] (neighbour3-feature) {};
            \node[feature, draw=orangeish, below=of neighbour2-occlusion, fill=black, text=white] (neighbour3-occlusion) {\faEyeSlash};
            \node[feature, draw=orangeish, below=of neighbour3-feature] (neighbour4-feature) {};
            \node[feature, draw=orangeish, below=of neighbour3-occlusion, fill=black, text=white] (neighbour4-occlusion) {\faEyeSlash};
            \node[feature, draw=orangeish, below=of neighbour4-feature] (neighbour5-feature) {};
            \node[feature, draw=orangeish, below=of neighbour4-occlusion, fill=white] (neighbour5-occlusion) {\faEye};
        \end{scope}
    \end{scope}
    \node[htext,left=of self-data,text=blueish] {Self};
    \node[htext,above=of self-data] {Features};
    \node[htext, left=0.8cm of neighbours-data,text=orangeish, rotate=90, anchor=center] {Neighbours};

    \begin{scope}[shift={(18,-3)}, local bounding box=masked-attention]
        \begin{scope}[shift={(0,0)},local bounding box=attention-linears]
            \node[attention] (Q) at (0,0) {Q};
            \node[attention, below=of Q] (K) {K};
            \node[attention, below=of K] (V) {V};
        \end{scope}
    \end{scope}
    \begin{scope}[on background layer,local bounding box=attention-background]
        \fill[fill=black!15, rounded corners=2pt] 
            ($(attention-linears.north west)+(-0.5, 0.5)$) -- 
            ($(attention-linears.north east)+( 0.5, 0.5)$) -- 
            ($(attention-linears.south east)+( 0.5,-0.5)$) -- 
            ($(attention-linears.south west)+(-0.5,-0.5)$) -- cycle;
    \end{scope}
    \node[htext,above=of attention-background] {Masked Attention};

    \begin{scope}[local bounding box=collector-all]
    \draw[line width=1pt] ($(all-data.north east)+(0.3,-0.2)$) to[bend left=10] ($(all-data.south east)+(0.3,0.2)$);
    \end{scope}
    \draw[-stealth,line width=1pt] ($(collector-all) + (0.15,0)$) -- (K);
    \draw[-stealth,line width=1pt] ($(collector-all) + (0.15,0)$) -- (V);

    \begin{scope}[local bounding box=collector-self]
    \draw[line width=1pt] ($(self-data.north east)+(0.6,-0.2)$) to[bend left=10] ($(self-data.south east)+(0.6,0.2)$);
    \end{scope}
    \draw[-stealth,line width=1pt] (collector-self) -- (Q);
    \draw[-stealth,line width=1pt] (collector-self) -- (Q);

    \end{tikzpicture}
}
\vspace*{-0.8cm}
    \caption{The point-wise attention mechanism.}\label{fig:point-wise attention}
\end{figure}

Finally, we also introduce a point-wise attention mechanism~\cite{Tang2023MVDiffusionarxiv}, leveraging the fact that the geometry of target object is known, and therefore correspondences between any two views (with respective occlusion maps) are readily available.
Every (latent) pixel bilinearly interpolates its correspondences in all other views (including the reference), and uses a small attention layer to attend to all of these, as seen in \cref{fig:point-wise attention}.
We mask the attention with the known occlusion map to ensure that only true correspondences contribute to this attention; each point also attends to its own features to stabilize training and avoid a fully masked-out attention.
For the lower resolution occlusion maps, we mean-pool the highest-resolution map and binarize this with a threshold of $0.2$, to smooth and dilate it at the same time.

\subsection{Classifier-Free Guidance and Multiple Guidance}

Classifier-free guidance~\cite{CFG} finds directions in prediction space that align better to the conditions.
To do so, it considers the differential between conditioned and unconditioned predictions, and exaggerates the final prediction towards this direction (effectively amplifying the effect of the conditioning signal).
Multiple guidance signals allow more fine-grained control over the applied differentials at the cost of more model evaluations~\cite{kant2024spad,brooks2023instructpix2pix,liu2022compositional}, by using the differentials \wrt{} unconditional~\cite{liu2022compositional} or partially-conditioned predictions~\cite{kant2024spad,brooks2023instructpix2pix}.
The former is most useful when the conditions are relatively independent.
Our approach, with a strong correlation between the conditions, benefits more from the latter approach.

We identify the \emph{text prompt} $\mathcal{T}$, \emph{geometry} $\mathcal{G}$, \emph{reference view} $\mathcal{R}$, and \emph{neighboring views} $\mathcal{V}$ as our conditions.
Geometry (as a surface normal map) is always used, because its absence does not lead to a meaningful prediction differential.
We use
\begin{equation}
\begin{split}
    \mathcal{D}(\vec{z}_{t+1} \vert \phi, \mathcal{G}, \phi, \phi) &+ \omega_\Omega (\mathcal{D}(\vec{z}_{t+1} \vert \mathcal{T}, \mathcal{G}, \phi, \mathcal{V}) - \mathcal{D}(\vec{z}_{t+1} \vert \phi, \mathcal{G}, \phi, \phi))\\ &+ \omega_\mathcal{V} (\mathcal{D}(\vec{z}_{t+1} \vert \mathcal{T}, \mathcal{G}, \mathcal{R}, \mathcal{V}) - \mathcal{D}(\vec{z}_{t+1} \vert \mathcal{T}, \mathcal{G}, \phi, \mathcal{V}))\label{eq:multiple guidance},
\end{split}
\end{equation}
where $\mathcal{D}$ denotes the diffusion model's prediction, and $\phi$ a dropped condition.
The text prompt is dropped out as usual for the base model (replacing it with an empty string in the case of SD2), while we ignore neighboring views by emptying the occlusion masks in the masked attention of \cref{fig:point-wise attention}.
Positional encodings and DINOv2 features are simplied zeroed, while the reference view process is replaced with gaussian noise and set to time step $T$, as in IM-3D~\cite{melas2024im3d}.

While image generation generally benefits from high CFG weight~\cite{CFG}, the guidance by a noise-free reference view is so strong that it dominates the output and introduces artifacts when amplified.
In inference, we therefore use $\omega_\Omega=7.5$ and $\omega_\mathcal{V}=3.5$ to prevent the reference view from dominating the output.

\subsection{Training Details}

We first train single-view Collaborative Control with Stable Diffusion 2.1 as base model~\cite{CollabControl} on $768\times{}768$ for a total of \nicenumber{200000} update steps with a batch size of $192$ and a learning rate of $3\times{}10^{-5}$, with frozen base model.
Subsequently, we train our multi-view model by adding the multi-view layers as in \cref{fig:architecture overview} and training for \nicenumber{100000} update steps with a batch size of $24$ and a learning rate of $3\times{}10^{-5}$ (each batch element now contains one reference view and $4$ neighboring views).
We have found it necessary to train both the single-view and multi-view PBR layers together to achieve best results --- the base model remains frozen throughout.
The multiple guidance signals are trained by randomly reverting to the less-conditioned terms in \cref{eq:multiple guidance} with a probability of $5\%$ each.
Furthermore, to safe-guard the quality of the single-view model, we also drop out all multi-view related conditioning with another $5\%$ probability.
Despite training only on a fixed number of views, inference is quite robust to the number of views.

\subsection{Fusion into a texture map}

The generated appearances need to be fused into a single texture map to be useful in graphics pipelines.
After the significant efforts to make the outputs multi-view consistent, we have found that a simple fusion approach is sufficient, as there are (nearly) no inconsistencies to resolve.
The main drawback we still face is that the fusion is not able to fill in areas that are occluded in all views (contrary to, say, an LRM~\cite{MetaTextureGen}).
For now, we simply distribute the generated views over the top hemisphere, and leave more complex fusion for future work.

We represent the object appearance in 3D space using tri-planes~\cite{Chan2022} and a small neural decoder that also consumes a positional embedding.
For faster optimization, we first fit this to low-resolution versions of the appearances using an MSE loss before fitting to the full-resolution appearances using both a perceptual LPIPS loss~\cite{LPIPS} (on randomly-drawn channel triplets~\cite{BRDFTripletLosses}) as well as an MSE loss --- please refer to \cref{lst:fusion algorithm} for the algorithm fusing the appearances into this global representation.

\begin{algorithm}[t]
	\caption{Fusion algorithm, using Kaolin Wisp's \texttt{models}~\cite{KaolinWisp} and \texttt{Pytorch}.}\label{lst:fusion algorithm}
	\begin{algorithmic}\small
        \State $\mathbf{T} \gets{}$ \texttt{TriplanarGrid(feature dim=32, log base resolution=6, num lods=4)}
        \State $\mathbf{P} \gets{}$ \texttt{PositionalEmbedder(num freq=8, input dim=3, max freq log2=8)}
        \State $\mathbf{D} \gets{}$ \texttt{BasicDecoder(output dim=8, activation=nn.SiLU, bias=True)}
        \State optimizer $\gets{}$ \texttt{Adam}(lr=$1\times{}10^{-2}$)
        \State appearances $\mathbf{A}$ of shape $N \times{} 8 \times{} 768 \times{} 768$
        \State world positions $\mathbf{W}$ of shape $N \times{} 3 \times{} 768 \times{} 768$
        \item[]
        \State $\mathbf{A}^\prime \gets \texttt{bilinear interpolate(}\mathbf{A}\texttt{, 1/32)}$
        \State $\mathbf{W}^\prime \gets \texttt{bilinear interpolate(}\mathbf{W}\texttt{, 1/32)}$
        \For {$iteration=1,\ldots,300$}
            \State modeled appearances $\tilde{\mathbf{A}}^\prime\gets \mathbf{D}(\mathbf{T}(\mathbf{W}^\prime), \mathbf{P}(\mathbf{W}^\prime))$
            \State loss $ \gets \texttt{MSE(}\tilde{\mathbf{A}}^\prime,\mathbf{A}^\prime\texttt{)}$
            \State \texttt{loss.backward()}
            \State \texttt{optimizer.step()}
        \EndFor
        \For {$iteration=1,\ldots,1000$}
            \State $\left\{\mathbf{A}_i\right\},\left\{\mathbf{W}_i\right\}\gets 64\ 32\times{}32$ patches from $\mathbf{A},\mathbf{W}$
            \State $\tilde{\mathbf{A}}_i \gets \mathbf{D}(\mathbf{T}(\mathbf{W}_i), \mathbf{P}(\mathbf{W}_i))$
            \State \texttt{MSE loss} $ \gets \sum\limits_i\texttt{MSE(}\tilde{\mathbf{A}}_i,\mathbf{A}_i\texttt{)}$
            \State channel triplet $\left\{c_j\right\} \gets \texttt{draw} 3 \texttt{from} 8$
            \State \texttt{LPIPS loss} $ \gets \sum\limits_{i,j}\texttt{LPIPS(}\tilde{\mathbf{A}}_{i,c_j},\mathbf{A}_{i,c_j}\texttt{)}$
            \State \texttt{loss} $ \gets \texttt{MSE loss} + 0.1 \times{} \texttt{LPIPS loss}$
            \State \texttt{loss.backward()}
            \State \texttt{optimizer.step()}
        \EndFor
	\end{algorithmic}
\end{algorithm}

Finally, we extract this global representation into a texture map by querying the global representation with the respective coordinates of the texture map (assuming a bijective mapping between the object and its texture).

\begin{figure}[t]
    \centering
    \includegraphics[width=0.11\linewidth,height=0.11\linewidth]{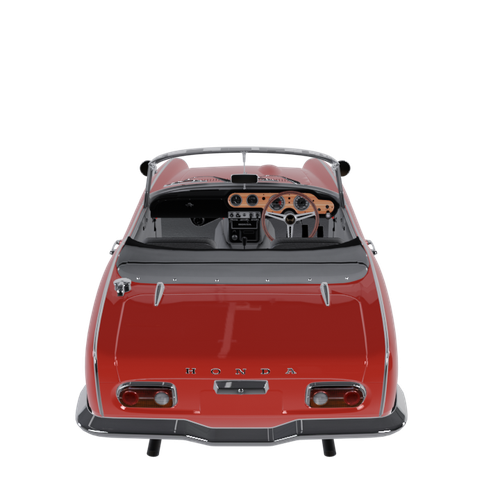}
    \includegraphics[width=0.11\linewidth,height=0.11\linewidth]{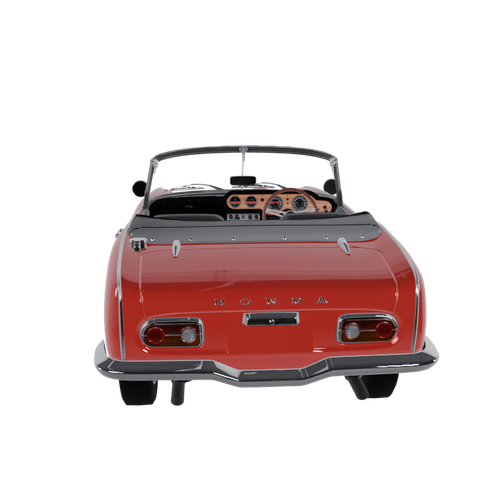}
    \includegraphics[width=0.11\linewidth,height=0.11\linewidth]{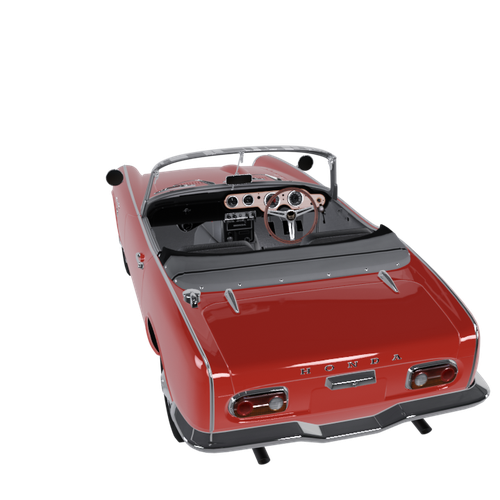}
    \includegraphics[width=0.11\linewidth,height=0.11\linewidth]{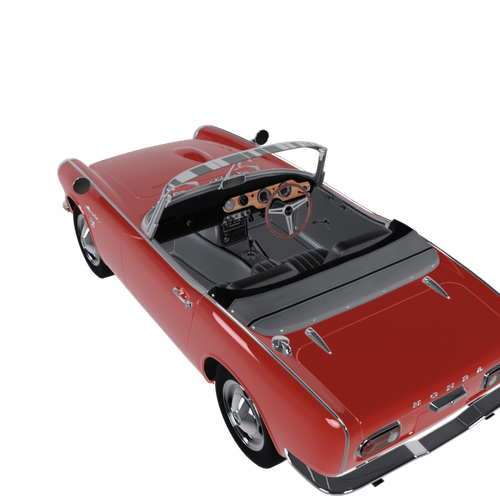}
    \includegraphics[width=0.11\linewidth,height=0.11\linewidth]{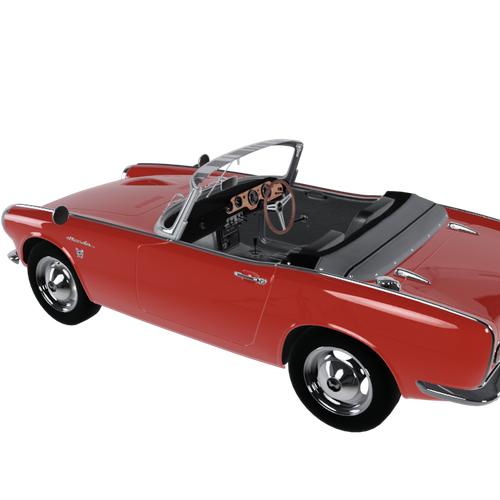}
    \includegraphics[width=0.11\linewidth,height=0.11\linewidth]{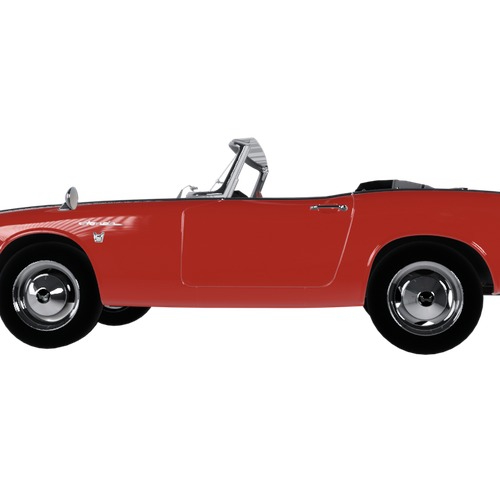}
    \includegraphics[width=0.11\linewidth,height=0.11\linewidth]{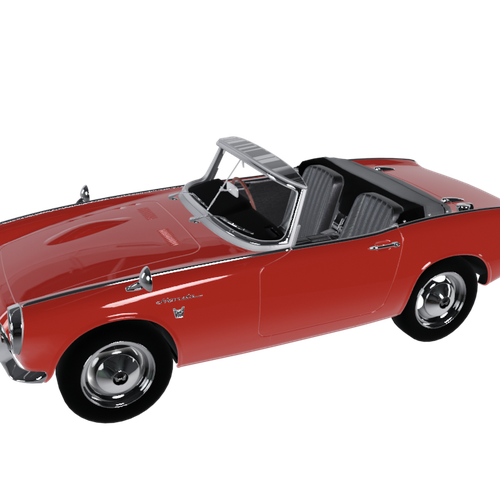}
    \includegraphics[width=0.11\linewidth,height=0.11\linewidth]{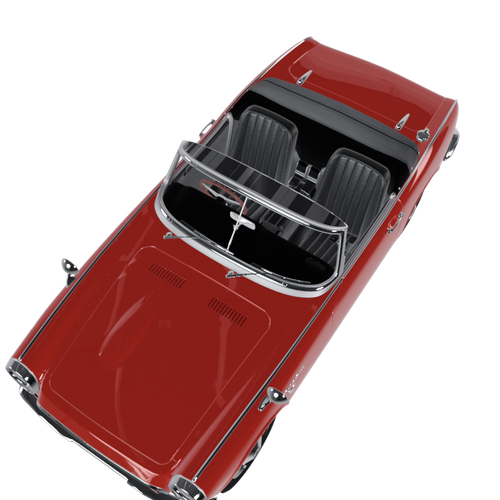}
    \includegraphics[width=0.11\linewidth,height=0.11\linewidth]{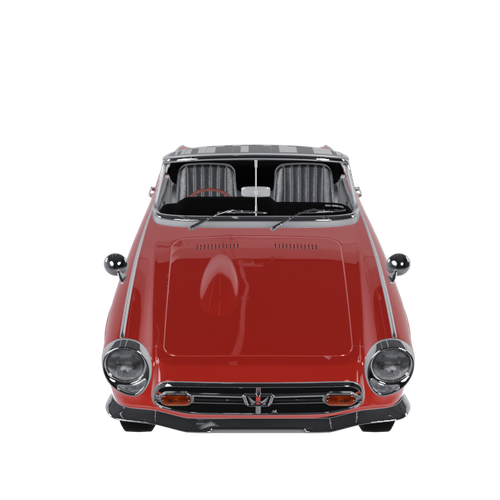}
    \includegraphics[width=0.11\linewidth,height=0.11\linewidth]{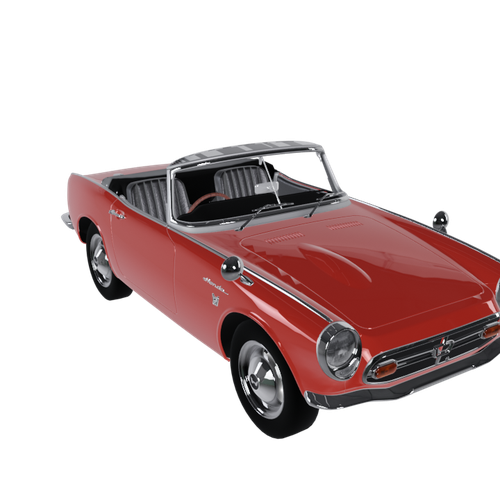}
    \includegraphics[width=0.11\linewidth,height=0.11\linewidth]{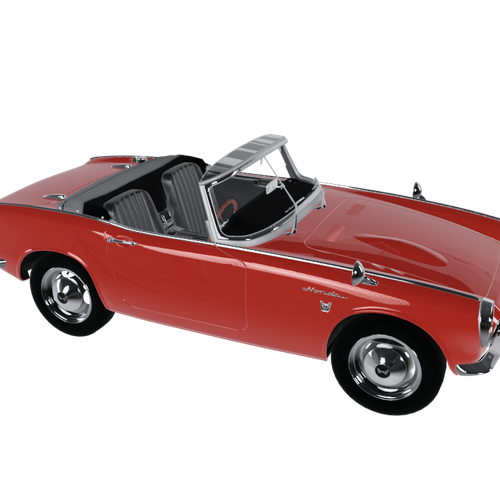}
    \includegraphics[width=0.11\linewidth,height=0.11\linewidth]{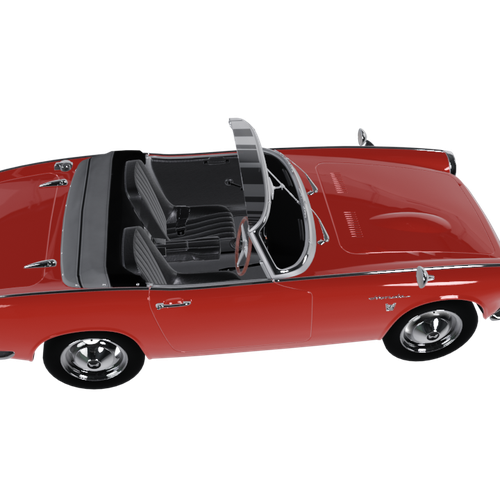}
    \includegraphics[width=0.11\linewidth,height=0.11\linewidth]{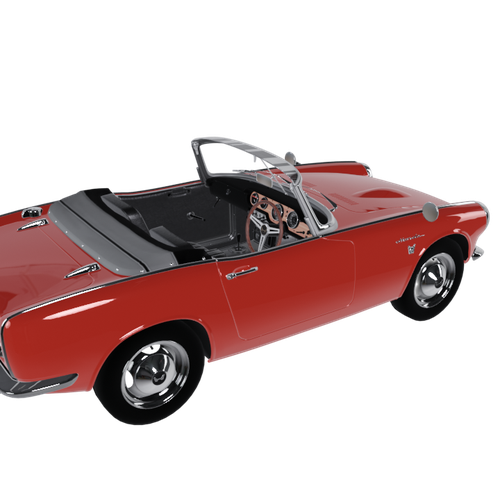}
    \includegraphics[width=0.11\linewidth,height=0.11\linewidth]{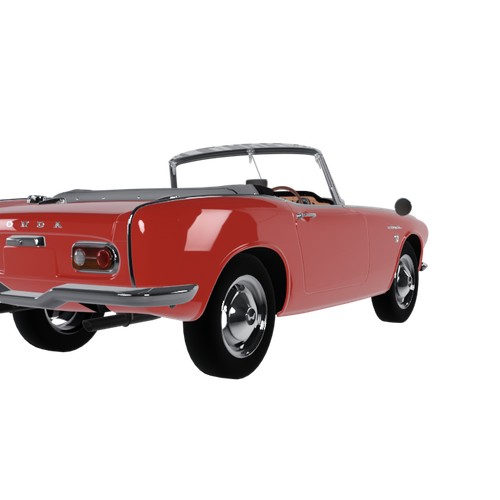}
    \includegraphics[width=0.11\linewidth,height=0.11\linewidth]{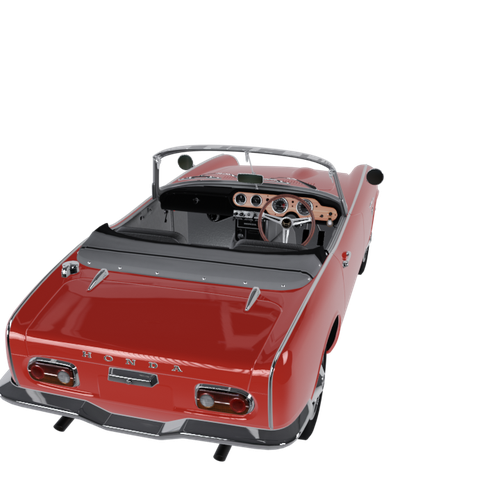}
    \includegraphics[width=0.11\linewidth,height=0.11\linewidth]{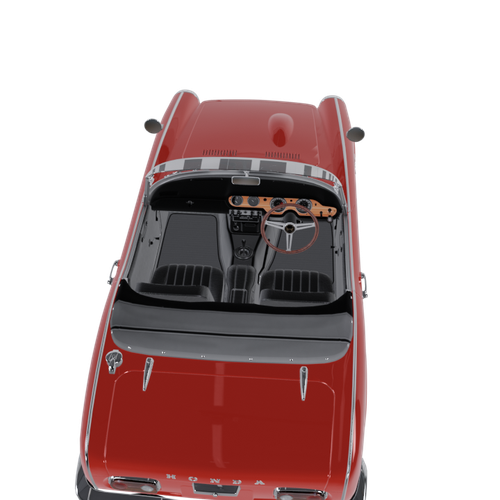}
    \caption{Example of the rendered views of an object~\cite{hondasketchfab}. The training images are rendered on a black background; the environment is shown here for illustration only.}\label{fig:render example}
\end{figure}

\section{Experiments and Results}\label{sec:results}

We now demonstrate the effectiveness of our proposed method, both in lifting the single-view generation to multi-view and in producing graphics-ready textures.
Given that the model is trained with dropped-out conditions of various types, we can re-use it to ablate the importance of those respective components.

\subsection{Dataset}

For training this model, both its single-view variant and its multi-view variant, we use Objaverse~\cite{Deitke2022Objaversearxiv}.
We remove objects without PBR textures or with non-spatially-varying textures, leaving roughly \nicenumber{300000} objects~\cite{CollabControl}.
The objects are rendered with a fixed (camera-corotated) environment map~\cite{StudioSmall08} from 16 viewpoints with varying elevation and equally spaced azimuths as shown in \cref{fig:render example}.
For evaluation, we re-use the held-out test set meshes as used in the Collaborative Control paper~\cite{CollabControl}, with the same ChatGPT-generated prompts per mesh~\cite{ChatGPT4}, from obvious to extremely unlikely.

\subsection{Ablating the new components}

We compare the full model to restricted versions of itself, where we ablate respectively the pixel-wise correspondence, its occlusion detection, the reference view hidden state attention and the DINOv2 cross attention.
For this section, to restrict the effect of the ablations, we only observe a single guidance signal
\begin{equation}
    \mathcal{D}(\vec{z}_{t+1} \vert \phi, \mathcal{G}, \phi, \phi) + \omega (\mathcal{D}(\vec{z}_{t+1} \vert \mathcal{T}, \mathcal{G}, \mathcal{R}, \mathcal{V}) - \mathcal{D}(\vec{z}_{t+1} \vert \phi, \mathcal{G}, \phi, \phi))
\end{equation}
instead of \cref{eq:multiple guidance}. We have found that this formulation performs best with a relatively low $\omega=1.5$.
For visual clarity, we cherry-pick one test mesh and prompt to visualize for each of the following ablations; the reference views were generated with the single-view model and the same prompt.
For compactness, we limit the visualizations to a few of the PBR maps as well as the final Blender-rendering of the textured mesh from the other side; the full set of outputs, including generated images, fused textures and rendered videos, is available in the supplementary material.

\clearpage
\noindent\begin{minipage}{\linewidth}%
    \centering
    \input{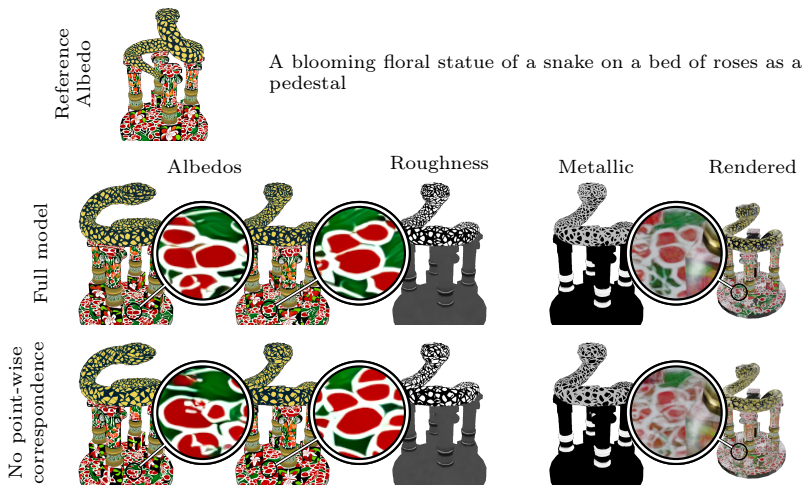}

    \captionof{figure}{Effect of removing the attention to pixel-wise correspondences from the multi-view layers in \cref{fig:multiview attention}. This is crucial for best results: without it, the generated views aren't pixel-consistent, resulting in ghosting after fusion into a mesh texture.}\label{fig:ablations pixelwise}
\end{minipage}

\noindent\begin{minipage}{\linewidth}%
    \centering
    \input{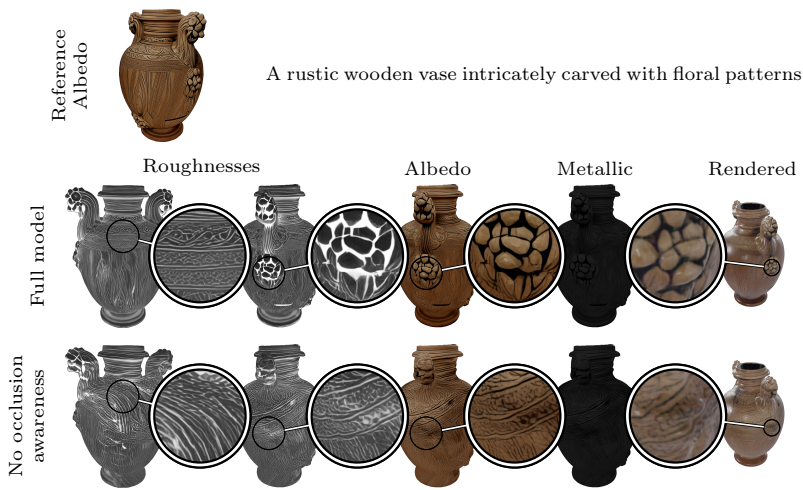}

    \captionof{figure}{Ignoring occlusions in the masked attention from \cref{fig:point-wise attention} results in features leaking towards back views, most noticeably from the reference view.}\label{fig:ablations occlusions}
\end{minipage}

\subsubsection{Pixel-wise correspondences} are, unsurprisingly, crucial for multi-view consistency.
As \cref{fig:ablations pixelwise} shows, omitting pixel-wise correspondence attention noticeably worsens fusion fitting loss, and consequently results in blurry, inconsistent textures in the final result.
Although the remaining components of the model manage to keep the style consistent and \eg{} prevent Janus artifacts, the lack of pixel-precise guidance has a significant impact on the result.
Occlusion detection is important to prevent features leaking through the object: as shown in \cref{fig:ablations occlusions} that is most visible through reference view features appearing in the back view.

\clearpage
\noindent\begin{minipage}{\linewidth}%
    \centering
    \input{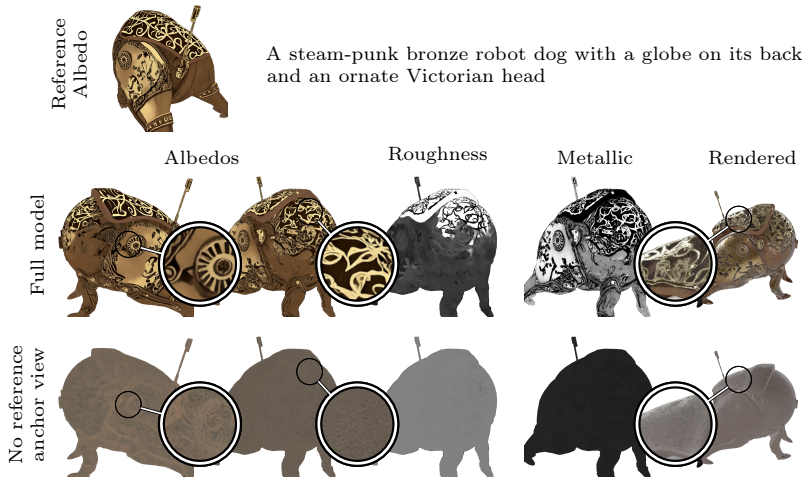}

    \captionof{figure}{Not attending to the reference view states in \cref{fig:multiview attention} shows clearly that the back views draw the majority of their content from this source.}\label{fig:ablations reference hidden states}
\end{minipage}

\noindent\begin{minipage}{\linewidth}%
    \centering
    \input{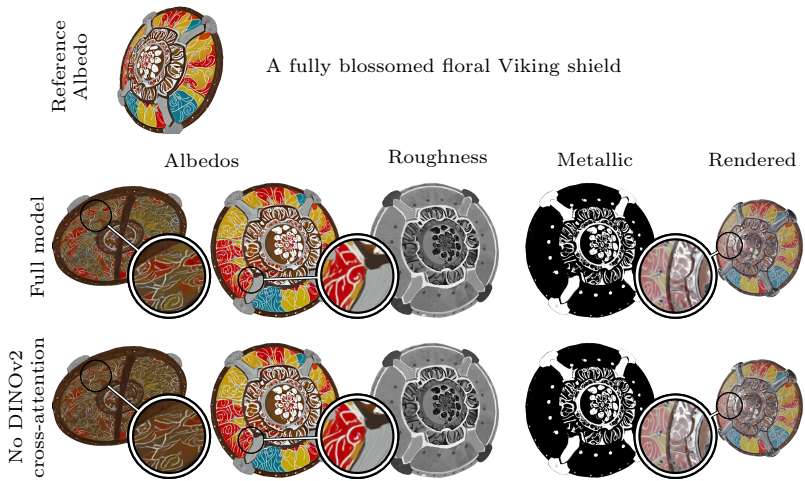}

    \captionof{figure}{The DINOv2 features of the reference view carry stylistic information. Although their impact is subtle, they help transporting some additional information to the back views. In the views that overlap significantly with the reference view, the absence of these stylistic features is not noticeable.}\label{fig:ablations reference DINO}
\end{minipage}

\subsubsection{Attention to the reference view or its DINOv2 features}

Omitting the reference image as batch element results in output that is multi-view consistent and matches the input image stylistically, but is not multi-view consistent with, as shown in \cref{fig:ablations reference hidden states}.
The cross-attention to the reference view's DINOv2 features~\cite{oquab2023dinov2}, promotes high-level coherency of the style across the views~\cite{jeong2023nvs,Zheng2023Free3Darxiv}, as seen in \cref{fig:ablations reference DINO} this is most important for the back views.

\begin{figure}[H]
    \centering
    \input{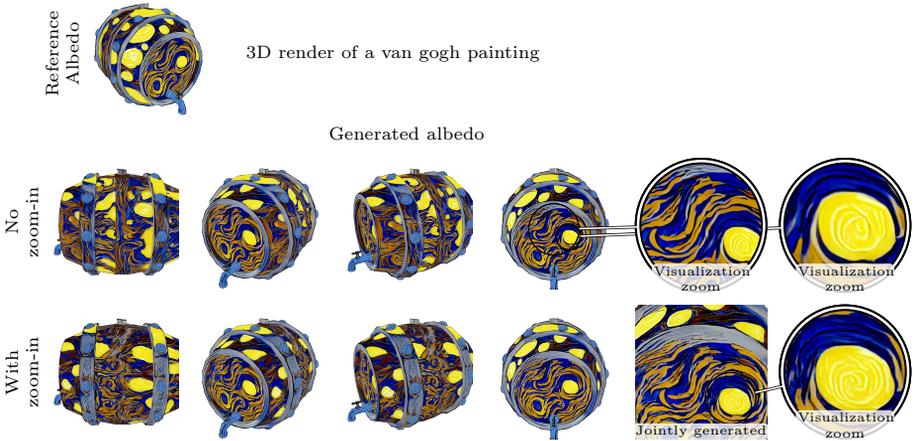}
    \caption{As the trained model shows to be robust to both the number of neighbors and the distance to the object, we can add zoom-ins to certain areas in order to boost the local texture quality, despite the restricted inference resolution of the model.}\label{fig:ablations views and scale}
\end{figure}\vspace*{-1cm}
\begin{figure}[H]
    \centering
    \input{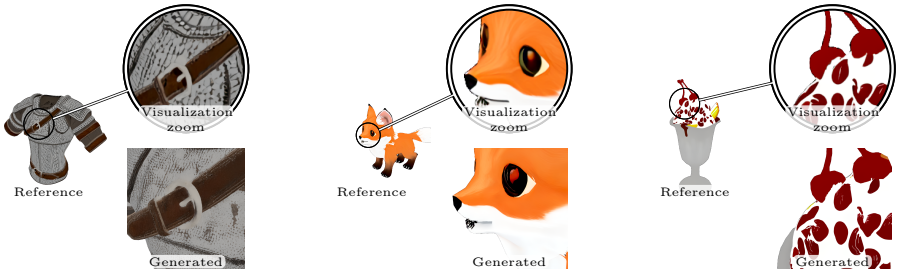}
    \caption{Here, we run the proposed method to generate only one novel view: a zoom-in on the reference image. Our model exhibits super-resolution capabilities, generating significantly higher-resolution output PBR images than the reference image. For these results, we have omitted the input text caption to the diffusion model.}\label{fig:superresolution gimmick}
\end{figure}

\subsection{Number of views and varying scale}

Considering the training details, which fix the number of neighbors to $4$ and the distance to the object to be constant, we find in practice that the approach is robust to either.
\Cref{fig:ablations views and scale} illustrates the effect of adding a zoomed-in view of part of the object.
This way, we generate high-resolution textures despite the base RGB image model's resolution of $768\times 768$.
Misusing this extremely powerful feature, \cref{fig:superresolution gimmick} indicates that this can even serve as a texture method, although we do not specifically evaluate this capability in this work, and we show only albedo images (all channels are available in the supplementary).

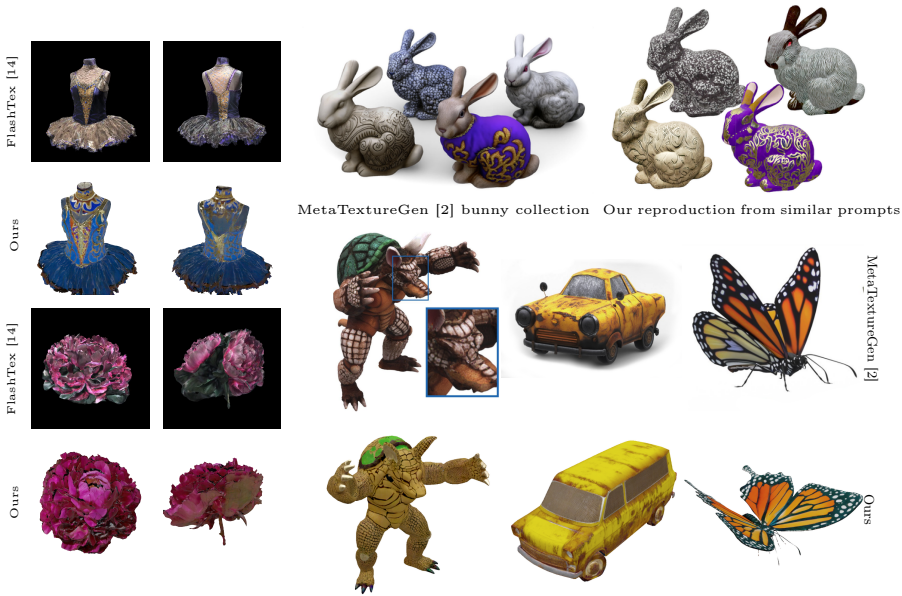
\begin{figure}[ht]
    \centering
    \begin{minipage}{0.3\linewidth}
        \centering
        \resizebox{\linewidth}{!}{\begin{tikzpicture}[
            node distance=0.2cm, 
            block/.style={inner sep=0mm, draw=none, transform shape},
            prompt/.style={inner sep=0cm, text width=0.57\linewidth, align=justify, transform shape},
            overlaytext/.style={inner sep=1pt,align=center,opacity=0.8,fill=white,text opacity=1.0,rounded corners=2pt,yshift=3pt,font=\tiny},
            label/.style={inner sep=0cm, transform shape, font=\tiny},
        ]
        \node[block] (flashtex1-1) at (0, 0) {\includegraphics[width=0.5\linewidth,height=0.5\linewidth]{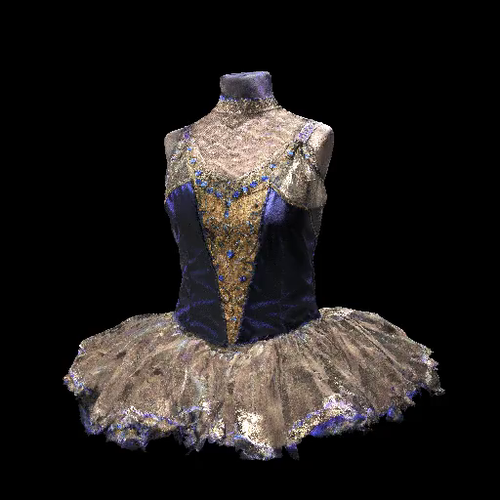}};
        \node[block, right=of flashtex1-1] (flashtex1-2) {\includegraphics[width=0.5\linewidth,height=0.5\linewidth]{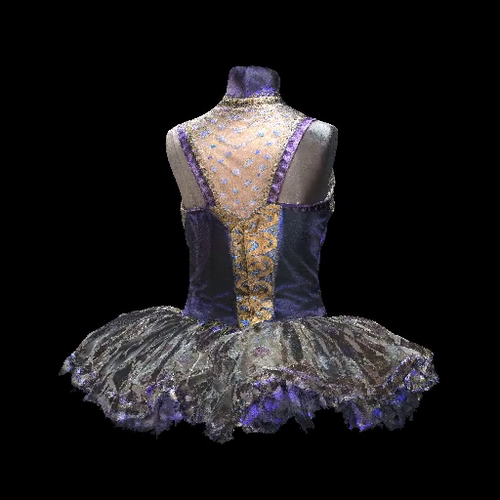}};

        \node[block, below=of flashtex1-1] (us1-1) {\includegraphics[width=0.5\linewidth,height=0.5\linewidth, clip]{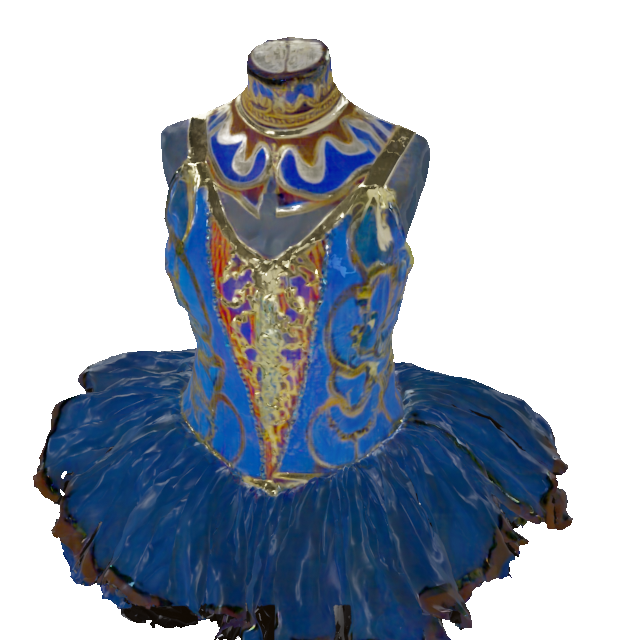}};
        \node[block, right=of us1-1] (us1-2) {\includegraphics[width=0.5\linewidth,height=0.5\linewidth, clip]{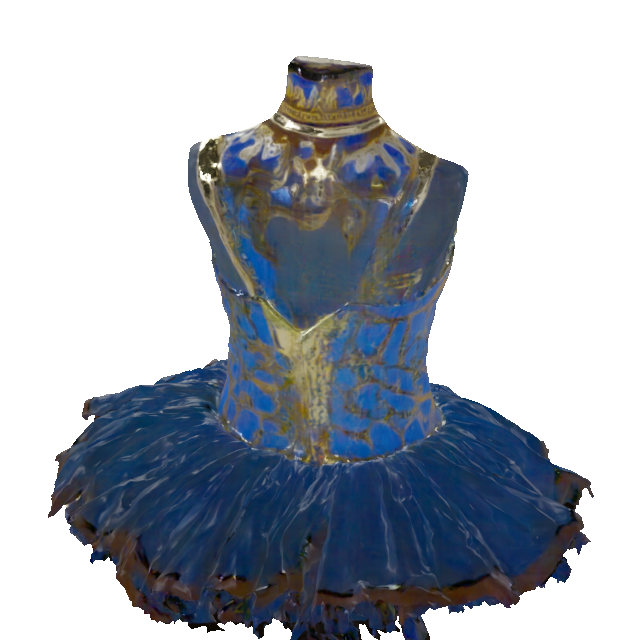}};

        \node[block, below=of us1-1] (flashtex2-1) {\includegraphics[width=0.5\linewidth,height=0.5\linewidth]{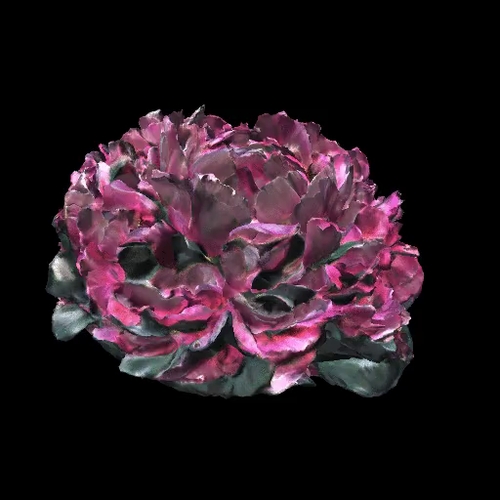}};
        \node[block, right=of flashtex2-1] (flashtex2-2) {\includegraphics[width=0.5\linewidth,height=0.5\linewidth]{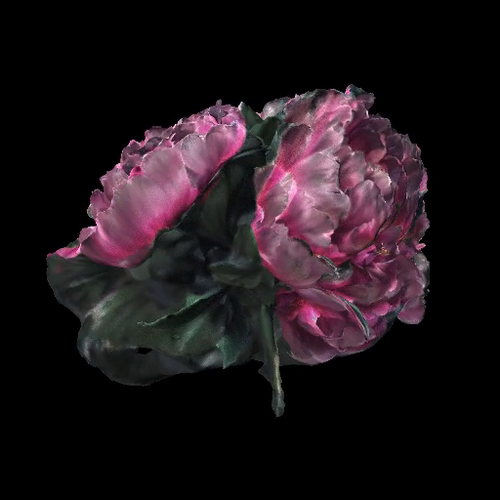}};

        \node[block, below=of flashtex2-1] (us2-1) {\includegraphics[width=0.5\linewidth,height=0.5\linewidth, clip]{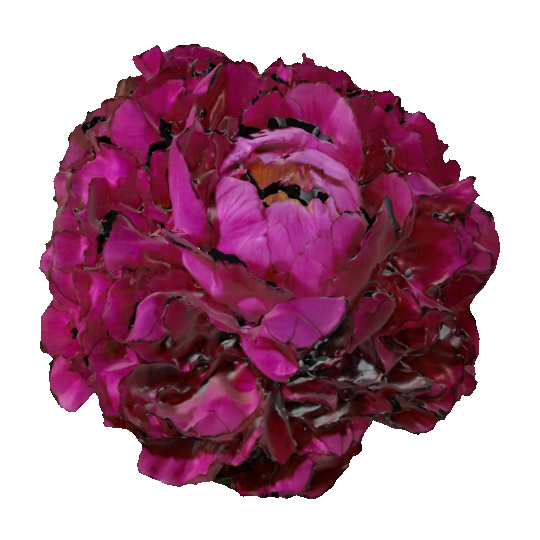}};
        \node[block, right=of us2-1] (us2-2) {\includegraphics[width=0.5\linewidth,height=0.5\linewidth, clip]{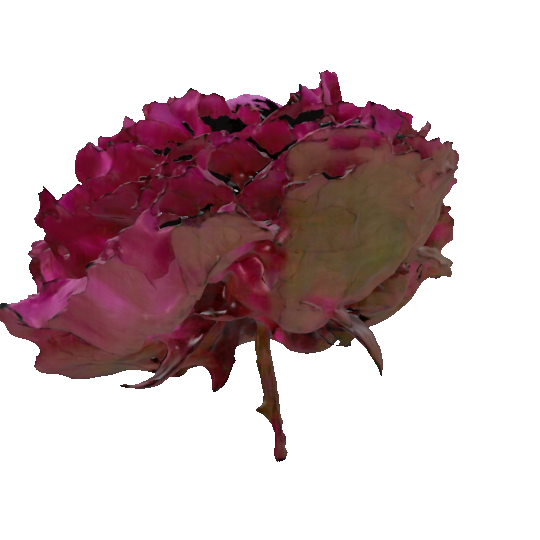}};

        \node[label, left=of flashtex1-1] {\rotatebox{90}{FlashTex~\cite{FlashTex}}};
        \node[label, left=of us1-1] {\rotatebox{90}{Ours}};
        \node[label, left=of flashtex2-1] {\rotatebox{90}{FlashTex~\cite{FlashTex}}};
        \node[label, left=of us2-1] {\rotatebox{90}{Ours}};
        \end{tikzpicture}}
    \end{minipage}
    \begin{minipage}{0.68\linewidth}\centering
        \begin{minipage}[b]{0.48\linewidth}
            \includegraphics[width=\linewidth]{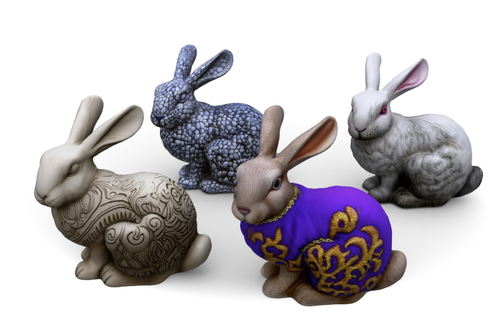}
            \tiny MetaTextureGen~\cite{MetaTextureGen} bunny collection
        \end{minipage}
        \begin{minipage}[b]{0.48\linewidth}
            \includegraphics[width=\linewidth]{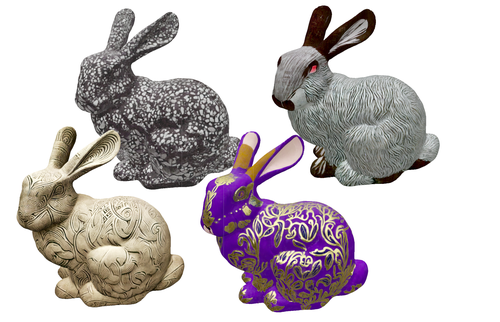}
            \tiny Our reproduction from similar prompts
        \end{minipage}
        
        \resizebox{0.9\linewidth}{!}{\begin{tikzpicture}[
            node distance=0.0cm, 
            block/.style={inner sep=0mm, draw=none, transform shape},
            prompt/.style={inner sep=0cm, text width=0.57\linewidth, align=justify, transform shape},
            overlaytext/.style={inner sep=1pt,align=center,opacity=0.8,fill=white,text opacity=1.0,rounded corners=2pt,yshift=3pt,font=\tiny},
            label/.style={inner sep=0cm, transform shape, font=\small},
        ]
        \node[block] (meta1-1) at (0, 0) {\includegraphics[width=0.5\linewidth]{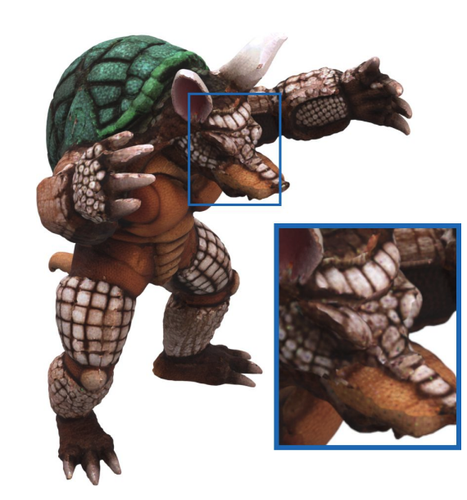}};
        \node[block, right=of meta1-1] (meta1-2) {\includegraphics[width=0.5\linewidth]{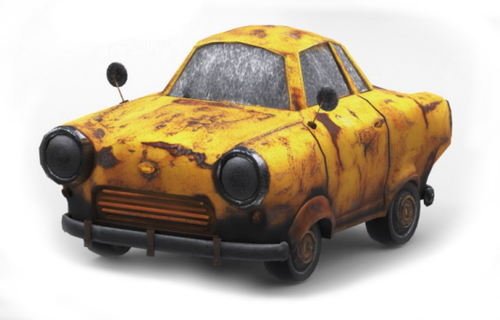}};
        \node[block, right=of meta1-2] (meta1-3) {\includegraphics[width=0.5\linewidth]{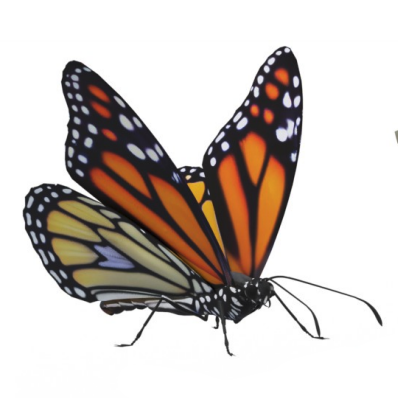}};

        \node[block, below=of meta1-1] (us1-1) {\includegraphics[width=0.5\linewidth]{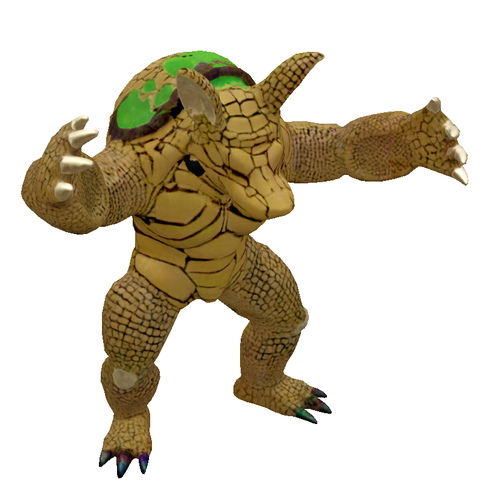}};
        \node[block, right=of us1-1] (us1-2) {\includegraphics[width=0.5\linewidth]{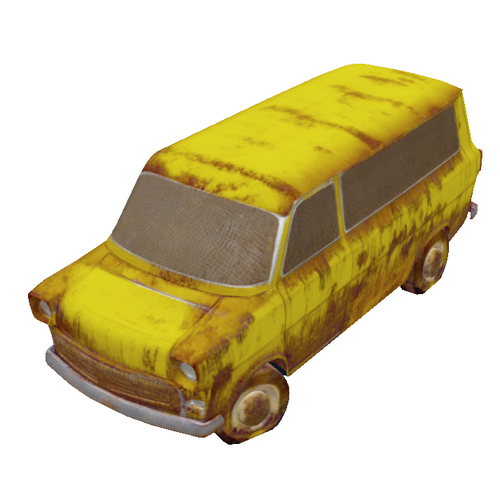}};
        \node[block, right=of us1-2] (us1-3) {\includegraphics[width=0.5\linewidth]{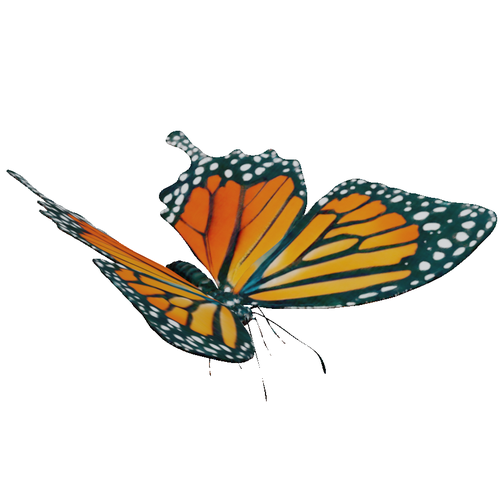}};

        \node[label, right=of meta1-3] {\rotatebox{-90}{MetaTextureGen~\cite{MetaTextureGen}}};
        \node[label, right=of us1-3] {\rotatebox{-90}{Ours}};
        \end{tikzpicture}}
    \end{minipage}
    
    \caption{Qualitative comparison of renders base on our outputs and those from FlashTex~\cite{FlashTex} and MetaTextureGen~\cite{MetaTextureGen} (images taken from their respective papers and supplementaries). Compared to FlashTex, which relies on SDS for PBR maps, we show cleaner appearances and an absence of Janus issues \eg{} on the back view of the peony.
    Our general output quality is high, but doesn't quite reach the same level of realism as MetaTextureGen.
    We attribute this in part to a relatively high CFG scale, and in part to the source dataset: while Objaverse has known drawbacks in terms of realism and quality, not much is known about the dataset used in MetaTextureGen.
    }\label{fig:qualitative comparison}
\end{figure}

\subsection{Qualitative comparisons to concurrent methods}
Finally, we offer a qualitative comparison to concurrent state-of-the-art Text-to-Texture methods. \Cref{fig:qualitative comparison} shows the results of our method compared to FlashTex~\cite{FlashTex} and MetaTextureGen~\cite{MetaTextureGen}, trying to reproduce similar meshes and prompts as the ones originally reported by the respective authors. MetaTextureGen is --- to our knowledge --- the only concurrent method to also directly output full PBR stacks.
However, the diffusion model only outputs albedo and ``specular'' images; an LRM~\cite{MetaTextureGen} produces the final PBR images based on these.

\begin{figure}
    \centering
    \input{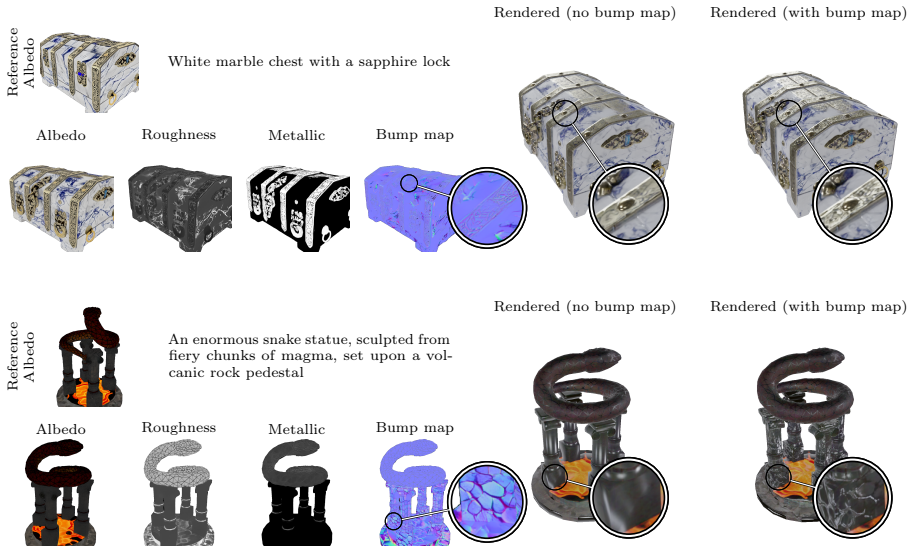}
    \caption{Our model introduces additional detail through the bump map whenever the geometry is too flat for the intended texture; in other cases, the detailed geometry is left intact.
    This allows artists to skip the time-consuming sculpting of detailed geometry and its subsequent baking out into the bump map.}\label{fig:showing off normal bump maps}
\end{figure}

Finally, we are not aware of any method that generates bump maps jointly with the other PBR images
High-resolution geometry is time-consuming to create and resource-intensive during rendering.
A typical workflow involves an artist sculpting detailed geometry before baking the details into a normal map and simplifying the underlying geometry.
Given the large number of objects in the Objaverse dataset~\cite{Deitke2022Objaversearxiv} that underwent this  process, our model has managed to learn how to reverse this process to some extent: it can generate high-resolution normal maps for the low-resolution meshes, as shown in \cref{fig:showing off normal bump maps}.

\section{Conclusion}\label{sec:conclusion}

We have shown how to extend the single-view PBR results to multi-view consistent predictions that can easily be fused into a mesh's texture.
As a result, we now have a complete pipeline from un-textured mesh to a graphics-compatible texture, which inherits the quality and generalizability of the underlying RGB image model thanks to the Collaborative Control scheme~\cite{CollabControl}.

The main drawback of our approach is that often, some areas of the texture remain unobserved in the generated views: we see a possible solution either in the use of an LRM, or by directly learning a diffusion model on the tri-plane space.
Additionally, practical use in commercial or otherwise public applications requires significant attention to ethical concerns around the sourcing of training data: both for the base RGB diffusion model and for the PBR model.

\section*{Acknowledgments}

This work was supported fully by Unity Technologies, without external funding.

\bibliographystyle{splncs04}
\bibliography{main,holo_library_manual,mark_library}

\begin{thebibliography}{10}
\providecommand{\url}[1]{\texttt{#1}}
\providecommand{\urlprefix}{URL }
\providecommand{\doi}[1]{https://doi.org/#1}

\bibitem{ChatGPT4}
Achiam, J., Adler, S., Agarwal, S., Ahmad, L., Akkaya, I., Aleman, F.L.,
  Almeida, D., Altenschmidt, J., Altman, S., Anadkat, S., et~al.: Gpt-4
  technical report. arXiv preprint arXiv:2303.08774  (2023)

\bibitem{MetaTextureGen}
Bensadoun, R., Kleiman, Y., Azuri, I., Harosh, O., Vedaldi, A., Neverova, N.,
  Gafni, O.: Meta 3d texturegen: Fast and consistent texture generation for 3d
  objects (2024)

\bibitem{brooks2023instructpix2pix}
Brooks, T., Holynski, A., Efros, A.A.: Instructpix2pix: Learning to follow
  image editing instructions. In: Proceedings of the IEEE/CVF Conference on
  Computer Vision and Pattern Recognition. pp. 18392--18402 (2023)

\bibitem{Burley2012}
Burley, B.: Physically based shading at disney. In: ACM Transactions on
  Graphics (SIGGRAPH) (2012)

\bibitem{cao2023texfusion}
Cao, T., Kreis, K., Fidler, S., Sharp, N., Yin, K.: Texfusion: Synthesizing 3d
  textures with text-guided image diffusion models. In: Proceedings of the
  IEEE/CVF International Conference on Computer Vision. pp. 4169--4181 (2023)

\bibitem{BRDFTripletLosses}
Chambon, T., Heitz, E., Belcour, L.: Passing multi-channel material textures to
  a 3-channel loss  (May 2021). \doi{10.1145/3450623.3464685},
  \url{http://arxiv.org/abs/2105.13012v1}

\bibitem{Chan2022}
Chan, E.R., Lin, C.Z., Chan, M.A., Nagano, K., Pan, B., Mello, S.D., Gallo, O.,
  Guibas, L., Tremblay, J., Khamis, S., Karras, T., Wetzstein, G.: Efficient
  geometry-aware {3D} generative adversarial networks. In: CVPR (2022)

\bibitem{chen2023text2tex}
Chen, D.Z., Siddiqui, Y., Lee, H.Y., Tulyakov, S., Nie{\ss}ner, M.: Text2tex:
  Text-driven texture synthesis via diffusion models. In: Proceedings of the
  IEEE/CVF International Conference on Computer Vision. pp. 18558--18568 (2023)

\bibitem{chen2023fantasia3d}
Chen, R., Chen, Y., Jiao, N., Jia, K.: Fantasia3d: Disentangling geometry and
  appearance for high-quality text-to-3d content creation. In: Proceedings of
  the IEEE/CVF international conference on computer vision. pp. 22246--22256
  (2023)

\bibitem{chen2024intrinsicanything}
Chen, X., Peng, S., Yang, D., Liu, Y., Pan, B., Lv, C., Zhou, X.:
  Intrinsicanything: Learning diffusion priors for inverse rendering under
  unknown illumination. arXiv preprint arXiv:2404.11593  (2024)

\bibitem{Chen2023Cascade-Zero123arxiv}
Chen, Y., Fang, J., Huang, Y., Yi, T., Zhang, X., Xie, L., Wang, X., Dai, W.,
  Xiong, H., Tian, Q.: Cascade-zero123: One image to highly consistent 3d with
  self-prompted nearby views  (Dec 2023),
  \url{http://arxiv.org/abs/2312.04424v1}

\bibitem{Cook1982}
Cook, R.L., Torrance, K.E.: A reflectance model for computer graphics. ACM
  Transactions on Graphics (ToG)  (1982)

\bibitem{Deitke2022Objaversearxiv}
Deitke, M., Schwenk, D., Salvador, J., Weihs, L., Michel, O., VanderBilt, E.,
  Schmidt, L., Ehsani, K., Kembhavi, A., Farhadi, A.: Objaverse: A universe of
  annotated 3d objects  (Dec 2022), \url{http://arxiv.org/abs/2212.08051v1}

\bibitem{FlashTex}
Deng, K., Omernick, T., Weiss, A., Ramanan, D., Zhu, J.Y., Zhou, T., Agrawala,
  M.: Flashtex: Fast relightable mesh texturing with lightcontrolnet (2024),
  \url{https://arxiv.org/abs/2402.13251}

\bibitem{ding2024bidiff}
Ding, L., Dong, S., Huang, Z., Wang, Z., Zhang, Y., Gong, K., Xu, D., Xue, T.:
  Text-to-3d generation with bidirectional diffusion using both 2d and 3d
  priors. In: Proceedings of the IEEE/CVF Conference on Computer Vision and
  Pattern Recognition. pp. 5115--5124 (2024)

\bibitem{esser2024SD3}
Esser, P., Kulal, S., Blattmann, A., Entezari, R., Müller, J., Saini, H.,
  Levi, Y., Lorenz, D., Sauer, A., Boesel, F., Podell, D., Dockhorn, T.,
  English, Z., Lacey, K., Goodwin, A., Marek, Y., Rombach, R.: Scaling
  rectified flow transformers for high-resolution image synthesis (2024),
  \url{https://arxiv.org/abs/2403.03206}

\bibitem{goodfellow2020generative}
Goodfellow, I., Pouget-Abadie, J., Mirza, M., Xu, B., Warde-Farley, D., Ozair,
  S., Courville, A., Bengio, Y.: Generative adversarial networks.
  Communications of the ACM  \textbf{63}(11),  139--144 (2020)

\bibitem{ho2020denoising}
Ho, J., Jain, A., Abbeel, P.: Denoising diffusion probabilistic models.
  Advances in neural information processing systems  \textbf{33},  6840--6851
  (2020)

\bibitem{CFG}
Ho, J., Salimans, T.: Classifier-free diffusion guidance. In: NeurIPS 2021
  Workshop on Deep Generative Models and Downstream Applications (2021)

\bibitem{Ho2022Classifier-Freearxiv}
Ho, J., Salimans, T.: Classifier-free diffusion guidance  (Jul 2022),
  \url{http://arxiv.org/abs/2207.12598v1}

\bibitem{hollein2024viewdiff}
H{\"o}llein, L., M{\"u}ller, N., Novotny, D., Tseng, H.Y., Richardt, C.,
  Zollh{\"o}fer, M., Nie{\ss}ner, M., et~al.: Viewdiff: 3d-consistent image
  generation with text-to-image models. In: Proceedings of the IEEE/CVF
  Conference on Computer Vision and Pattern Recognition. pp. 5043--5052 (2024)

\bibitem{hu2024mvd}
Hu, H., Zhou, Z., Jampani, V., Tulsiani, S.: Mvd-fusion: Single-view 3d via
  depth-consistent multi-view generation. In: Proceedings of the IEEE/CVF
  Conference on Computer Vision and Pattern Recognition. pp. 9698--9707 (2024)

\bibitem{Huang2023EpiDiffarxiv}
Huang, Z., Wen, H., Dong, J., Wang, Y., Li, Y., Chen, X., Cao, Y.P., Liang, D.,
  Qiao, Y., Dai, B., Sheng, L.: Epidiff: Enhancing multi-view synthesis via
  localized epipolar-constrained diffusion  (Dec 2023),
  \url{http://arxiv.org/abs/2312.06725v1}

\bibitem{sphericalshipalbedo}
IrisProcess: Spherical ship sketchfab model,
  \url{https://sketchfab.com/3d-models/spherical-ship-a1bed2ab95544638ba24589679a0e898}
  [Accessed: July 23rd, 2024]

\bibitem{jeong2023nvs}
Jeong, Y., Lee, J., Kim, C., Cho, M., Lee, D.: Nvs-adapter: Plug-and-play novel
  view synthesis from a single image. arXiv preprint arXiv:2312.07315  (2023)

\bibitem{kant2024spad}
Kant, Y., Wu, Z., Vasilkovsky, M., Qian, G., Ren, J., Guler, R.A., Ghanem, B.,
  Tulyakov, S., Gilitschenski, I., Siarohin, A.: Spad : Spatially aware
  multiview diffusers (2024), \url{https://arxiv.org/abs/2402.05235}

\bibitem{karras2017progressive}
Karras, T., Aila, T., Laine, S., Lehtinen, J.: Progressive growing of gans for
  improved quality, stability, and variation. arXiv preprint arXiv:1710.10196
  (2017)

\bibitem{karras2021alias}
Karras, T., Aittala, M., Laine, S., H{\"a}rk{\"o}nen, E., Hellsten, J.,
  Lehtinen, J., Aila, T.: Alias-free generative adversarial networks. Advances
  in Neural Information Processing Systems  \textbf{34},  852--863 (2021)

\bibitem{karras2019style}
Karras, T., Laine, S., Aila, T.: A style-based generator architecture for
  generative adversarial networks. In: Proceedings of the IEEE/CVF conference
  on computer vision and pattern recognition. pp. 4401--4410 (2019)

\bibitem{karras2020analyzing}
Karras, T., Laine, S., Aittala, M., Hellsten, J., Lehtinen, J., Aila, T.:
  Analyzing and improving the image quality of stylegan. In: Proceedings of the
  IEEE/CVF conference on computer vision and pattern recognition. pp.
  8110--8119 (2020)

\bibitem{kim2024referenceimage}
Kim, S., Shi, Y., Li, K., Cho, M., Wang, P.: Multi-view image prompted
  multi-view diffusion for improved 3d generation (2024),
  \url{https://arxiv.org/abs/2404.17419}

\bibitem{knodt2023consistent}
Knodt, J., Gao, X.: Consistent mesh diffusion. arXiv preprint arXiv:2312.00971
  (2023)

\bibitem{le2023euclidreamer}
Le, C., Hetang, C., Cao, A., He, Y.: Euclidreamer: Fast and high-quality
  texturing for 3d models with stable diffusion depth. arXiv preprint
  arXiv:2311.15573  (2023)

\bibitem{li2024era3d}
Li, P., Liu, Y., Long, X., Zhang, F., Lin, C., Li, M., Qi, X., Zhang, S., Luo,
  W., Tan, P., et~al.: Era3d: High-resolution multiview diffusion using
  efficient row-wise attention. arXiv preprint arXiv:2405.11616  (2024)

\bibitem{Instant-3D}
Li, S., Li, C., Zhu, W., Yu, B., Zhao, Y., Wan, C., You, H., Shi, H., Lin, Y.:
  Instant-3d: Instant neural radiance field training towards on-device ar/vr 3d
  reconstruction. In: Proceedings of the 50th Annual International Symposium on
  Computer Architecture. pp. 1--13 (2023)

\bibitem{liu2022compositional}
Liu, N., Li, S., Du, Y., Torralba, A., Tenenbaum, J.B.: Compositional visual
  generation with composable diffusion models. In: European Conference on
  Computer Vision. pp. 423--439. Springer (2022)

\bibitem{liu2023syncdreamer}
Liu, Y., Lin, C., Zeng, Z., Long, X., Liu, L., Komura, T., Wang, W.:
  Syncdreamer: Generating multiview-consistent images from a single-view image.
  arXiv preprint arXiv:2309.03453  (2023)

\bibitem{SyncMVD}
Liu, Y., Xie, M., Liu, H., Wong, T.T.: Text-guided texturing by synchronized
  multi-view diffusion (2023), \url{https://arxiv.org/abs/2311.12891}

\bibitem{liu2023unidream}
Liu, Z., Li, Y., Lin, Y., Yu, X., Peng, S., Cao, Y.P., Qi, X., Huang, X.,
  Liang, D., Ouyang, W.: Unidream: Unifying diffusion priors for relightable
  text-to-3d generation. arXiv preprint arXiv:2312.08754  (2023)

\bibitem{long2023wonder3d}
Long, X., Guo, Y.C., Lin, C., Liu, Y., Dou, Z., Liu, L., Ma, Y., Zhang, S.H.,
  Habermann, M., Theobalt, C., et~al.: Wonder3d: Single image to 3d using
  cross-domain diffusion. In: Proceedings of the IEEE/CVF Conference on
  Computer Vision and Pattern Recognition. pp. 9970--9980 (2024)

\bibitem{Lu2023Direct2.5arxiv}
Lu, Y., Zhang, J., Li, S., Fang, T., McKinnon, D., Tsin, Y., Quan, L., Cao, X.,
  Yao, Y.: Direct2.5: Diverse text-to-3d generation via multi-view 2.5d
  diffusion  (Nov 2023), \url{http://arxiv.org/abs/2311.15980v1}

\bibitem{StudioSmall08}
Majboroda, S.: \texttt{Studio small 08} hdri environment map,
  \url{https://polyhaven.com/a/studio_small_08} [Accessed: July 21st, 2024]

\bibitem{melas2024im3d}
Melas-Kyriazi, L., Laina, I., Rupprecht, C., Neverova, N., Vedaldi, A., Gafni,
  O., Kokkinos, F.: Im-3d: Iterative multiview diffusion and reconstruction for
  high-quality 3d generation. arXiv preprint arXiv:2402.08682  (2024)

\bibitem{oquab2023dinov2}
Oquab, M., Darcet, T., Moutakanni, T., Vo, H., Szafraniec, M., Khalidov, V.,
  Fernandez, P., Haziza, D., Massa, F., El-Nouby, A., et~al.: Dinov2: Learning
  robust visual features without supervision. arXiv preprint arXiv:2304.07193
  (2023)

\bibitem{podell2023SDXL}
Podell, D., English, Z., Lacey, K., Blattmann, A., Dockhorn, T., Müller, J.,
  Penna, J., Rombach, R.: Sdxl: Improving latent diffusion models for
  high-resolution image synthesis (2023),
  \url{https://arxiv.org/abs/2307.01952}

\bibitem{poole2022dreamfusion}
Poole, B., Jain, A., Barron, J.T., Mildenhall, B.: Dreamfusion: Text-to-3d
  using 2d diffusion. In: The Eleventh International Conference on Learning
  Representations (2023)

\bibitem{Radford2021LearningTV}
Radford, A., Kim, J.W., Hallacy, C., Ramesh, A., Goh, G., Agarwal, S., Sastry,
  G., Askell, A., Mishkin, P., Clark, J., Krueger, G., Sutskever, I.: Learning
  transferable visual models from natural language supervision. In:
  International Conference on Machine Learning (2021),
  \url{https://api.semanticscholar.org/CorpusID:231591445}

\bibitem{TEXTure}
Richardson, E., Metzer, G., Alaluf, Y., Giryes, R., Cohen-Or, D.: Texture:
  Text-guided texturing of 3d shapes (2023),
  \url{https://arxiv.org/abs/2302.01721}

\bibitem{rombach2022high}
Rombach, R., Blattmann, A., Lorenz, D., Esser, P., Ommer, B.: High-resolution
  image synthesis with latent diffusion models. In: Proceedings of the IEEE/CVF
  conference on computer vision and pattern recognition. pp. 10684--10695
  (2022)

\bibitem{Sartor2023Matfusion}
Sartor, S., Peers, P.: Matfusion: A generative diffusion model for svbrdf
  capture. In: SIGGRAPH Asia 2023 Conference Papers. SA ’23, ACM (Dec 2023).
  \doi{10.1145/3610548.3618194},
  \url{http://dx.doi.org/10.1145/3610548.3618194}

\bibitem{shi2023zero123++}
Shi, R., Chen, H., Zhang, Z., Liu, M., Xu, C., Wei, X., Chen, L., Zeng, C., Su,
  H.: Zero123++: a single image to consistent multi-view diffusion base model.
  arXiv preprint arXiv:2310.15110  (2023)

\bibitem{shi2023mvdream}
Shi, Y., Wang, P., Ye, J., Mai, L., Li, K., Yang, X.: Mvdream: Multi-view
  diffusion for 3d generation. In: The Twelfth International Conference on
  Learning Representations (2024)

\bibitem{LFNPlucker}
Sitzmann, V., Rezchikov, S., Freeman, W.T., Tenenbaum, J.B., Durand, F.: Light
  field networks: Neural scene representations with single-evaluation rendering
  (2022), \url{https://arxiv.org/abs/2106.02634}

\bibitem{sohl2015deep}
Sohl-Dickstein, J., Weiss, E., Maheswaranathan, N., Ganguli, S.: Deep
  unsupervised learning using nonequilibrium thermodynamics. In: International
  conference on machine learning. pp. 2256--2265. PMLR (2015)

\bibitem{song2020denoising}
Song, J., Meng, C., Ermon, S.: Denoising diffusion implicit models. In:
  International Conference on Learning Representations (2020)

\bibitem{KaolinWisp}
Takikawa, T., Perel, O., Tsang, C.F., Loop, C., Litalien, J., Tremblay, J.,
  Fidler, S., Shugrina, M.: Kaolin wisp: A pytorch library and engine for
  neural fields research. \url{https://github.com/NVIDIAGameWorks/kaolin-wisp}
  (2022)

\bibitem{InTeX}
Tang, J., Lu, R., Chen, X., Wen, X., Zeng, G., Liu, Z.: Intex: Interactive
  text-to-texture synthesis via unified depth-aware inpainting. arXiv preprint
  arXiv:2403.11878  (2024)

\bibitem{Tang2023MVDiffusionarxiv}
Tang, S., Zhang, F., Chen, J., Wang, P., Furukawa, Y.: Mvdiffusion: Enabling
  holistic multi-view image generation with correspondence-aware diffusion
  (Jul 2023), \url{http://arxiv.org/abs/2307.01097v7}

\bibitem{PBRexamplesketchfab}
Teczan, K.: Free game character - ancient sketchfab model,
  \url{https://sketchfab.com/3d-models/free-game-character-ancient-19e18c8e8d5941deb3f9dd0a2a349b56}
  [Accessed: July 23rd, 2024]

\bibitem{CollabControl}
Vainer, S., Boss, M., Parger, M., Kutsy, K., De~Nigris, D., Rowles, C., Perony,
  N., Donn{\'e}, S.: Collaborative control for geometry-conditioned pbr image
  generation (2024)

\bibitem{MatFuse}
Vecchio, G., Sortino, R., Palazzo, S., Spampinato, C.: Matfuse: Controllable
  material generation with diffusion models (2024),
  \url{https://arxiv.org/abs/2308.11408}

\bibitem{SV3D}
Voleti, V., Yao, C.H., Boss, M., Letts, A., Pankratz, D., Tochilkin, D.,
  Laforte, C., Rombach, R., Jampani, V.: Sv3d: Novel multi-view synthesis and
  3d generation from a single image using latent video diffusion (2024),
  \url{https://arxiv.org/abs/2403.12008}

\bibitem{wang2023mvdd}
Wang, Z., Xu, Q., Tan, F., Chai, M., Liu, S., Pandey, R., Fanello, S., Kadambi,
  A., Zhang, Y.: Mvdd: Multi-view depth diffusion models. arXiv preprint
  arXiv:2312.04875  (2023)

\bibitem{wen2024ouroboros3d}
Wen, H., Huang, Z., Wang, Y., Chen, X., Qiao, Y., Sheng, L.: Ouroboros3d:
  Image-to-3d generation via 3d-aware recursive diffusion. arXiv preprint
  arXiv:2406.03184  (2024)

\bibitem{woo2024harmonyview}
Woo, S., Park, B., Go, H., Kim, J.Y., Kim, C.: Harmonyview: Harmonizing
  consistency and diversity in one-image-to-3d. In: Proceedings of the IEEE/CVF
  Conference on Computer Vision and Pattern Recognition. pp. 10574--10584
  (2024)

\bibitem{wu2024unique3d}
Wu, K., Liu, F., Cai, Z., Yan, R., Wang, H., Hu, Y., Duan, Y., Ma, K.:
  Unique3d: High-quality and efficient 3d mesh generation from a single image
  (2024), \url{https://arxiv.org/abs/2405.20343}

\bibitem{wu2023hyperdreamer}
Wu, T., Li, Z., Yang, S., Zhang, P., Pan, X., Wang, J., Lin, D., Liu, Z.:
  Hyperdreamer: Hyper-realistic 3d content generation and editing from a single
  image. In: SIGGRAPH Asia 2023 Conference Papers. pp. 1--10 (2023)

\bibitem{xu2023matlaber}
Xu, X., Lyu, Z., Pan, X., Dai, B.: Matlaber: Material-aware text-to-3d via
  latent brdf auto-encoder. arXiv preprint arXiv:2308.09278  (2023)

\bibitem{hondasketchfab}
Yakpower: Honda s800 sketchfab model,
  \url{https://sketchfab.com/3d-models/honda-s800-fa6c6113f1e34d9baaff80b5b586a25d}
  [Accessed: July 23rd, 2024]

\bibitem{yang2024consistnet}
Yang, J., Cheng, Z., Duan, Y., Ji, P., Li, H.: Consistnet: Enforcing 3d
  consistency for multi-view images diffusion. In: Proceedings of the IEEE/CVF
  Conference on Computer Vision and Pattern Recognition. pp. 7079--7088 (2024)

\bibitem{ye2024consistent1to3consistentimage3d}
Ye, J., Wang, P., Li, K., Shi, Y., Wang, H.: Consistent-1-to-3: Consistent
  image to 3d view synthesis via geometry-aware diffusion models (2024),
  \url{https://arxiv.org/abs/2310.03020}

\bibitem{yeh2024texturedreamer}
Yeh, Y.Y., Huang, J.B., Kim, C., Xiao, L., Nguyen-Phuoc, T., Khan, N., Zhang,
  C., Chandraker, M., Marshall, C.S., Dong, Z., et~al.: Texturedreamer:
  Image-guided texture synthesis through geometry-aware diffusion. In:
  Proceedings of the IEEE/CVF Conference on Computer Vision and Pattern
  Recognition. pp. 4304--4314 (2024)

\bibitem{youwang2023paint}
Youwang, K., Oh, T.H., Pons-Moll, G.: Paint-it: Text-to-texture synthesis via
  deep convolutional texture map optimization and physically-based rendering.
  In: Proceedings of the IEEE/CVF Conference on Computer Vision and Pattern
  Recognition. pp. 4347--4356 (2024)

\bibitem{zeng2023paint3d}
Zeng, X., Chen, X., Qi, Z., Liu, W., Zhao, Z., Wang, Z., Fu, B., Liu, Y., Yu,
  G.: Paint3d: Paint anything 3d with lighting-less texture diffusion models.
  In: Proceedings of the IEEE/CVF Conference on Computer Vision and Pattern
  Recognition. pp. 4252--4262 (2024)

\bibitem{TexPainter}
Zhang, H., Pan, Z., Zhang, C., Zhu, L., Gao, X.: Texpainter: Generative mesh
  texturing with multi-view consistency (2024),
  \url{https://arxiv.org/abs/2406.18539}

\bibitem{zhang2023repaint123}
Zhang, J., Tang, Z., Pang, Y., Cheng, X., Jin, P., Wei, Y., Yu, W., Ning, M.,
  Yuan, L.: Repaint123: Fast and high-quality one image to 3d generation with
  progressive controllable 2d repainting. arXiv preprint arXiv:2312.13271
  (2023)

\bibitem{LPIPS}
Zhang, R., Isola, P., Efros, A.A., Shechtman, E., Wang, O.: The unreasonable
  effectiveness of deep features as a perceptual metric. In: CVPR (2018)

\bibitem{Zheng2023Free3Darxiv}
Zheng, C., Vedaldi, A.: Free3d: Consistent novel view synthesis without 3d
  representation  (Dec 2023), \url{http://arxiv.org/abs/2312.04551v1}

\end{thebibliography}
\end{document}